\documentclass[format=acmsmall]{acmart}
\usepackage{amsmath,amsfonts}
\usepackage{algorithmic}
\usepackage{algorithm}
\usepackage{array}
\usepackage{subfigure}
\usepackage{textcomp}
\usepackage{stfloats}
\usepackage{url}
\usepackage{verbatim}
\usepackage{graphicx}
\usepackage{multirow}
\usepackage{color}
\usepackage{arydshln}
\usepackage{booktabs}
\usepackage{bbding}

\AtBeginDocument{%
  \providecommand\BibTeX{{%
    \normalfont B\kern-0.5em{\scshape i\kern-0.25em b}\kern-0.8em\TeX}}}

\setcopyright{acmlicensed}
\acmJournal{TOMM}
\acmYear{2023} \acmVolume{1} \acmNumber{1} \acmArticle{1}
\acmMonth{1} \acmPrice{15.00}\acmDOI{10.1145/3631357}

\begin{document}
\title{Subjective and Objective Quality Assessment for in-the-Wild Computer Graphics Images}
\author{Zicheng Zhang}
\email{zzc1998@sjtu.edu.cn}
\author{Wei Sun*}
\email{sunguwei@sjtu.edu.cn}
\author{Yingjie Zhou}
\email{zyj2000@sjtu.edu.cn}
\author{Jun Jia}
\email{jiajun0302@sjtu.edu.cn}
\author{Zhichao Zhang}
\email{liquortect@sjtu.edu.cn}
\affiliation{%
  \institution{Shanghai Jiao Tong University}
  \country{China}
}
\author{Jing Liu}
\email{jliu\_tju@tju.edu.cn}
\affiliation{%
  \institution{Tianjin University}
  \country{China}
}
\author{Xiongkuo Min}
\email{minxiongkuo@sjtu.edu.cn}
\author{Guangtao Zhai*}
\email{zhaiguangtao@sjtu.edu.cn}
\affiliation{%
  \institution{Shanghai Jiao Tong University}
  \country{China}
}
\renewcommand{\shortauthors}{Zhang et al.}

\begin{abstract}
Computer graphics images (CGIs) are artificially generated by means of computer programs and are widely perceived under various scenarios, such as games, streaming media, etc. In practice, the quality of CGIs consistently suffers from poor rendering during production, inevitable compression artifacts during the transmission of multimedia applications, and low aesthetic quality resulting from poor composition and design. However, few works have been dedicated to dealing with the challenge of computer graphics image quality assessment (CGIQA). Most image quality assessment (IQA) metrics are developed for natural scene images (NSIs) and validated on databases consisting of NSIs with synthetic distortions, which are not suitable for in-the-wild CGIs. To bridge the gap between evaluating the quality of NSIs and CGIs, we construct a large-scale in-the-wild CGIQA database consisting of 6,000 CGIs (CGIQA-6k) and carry out the subjective experiment in a well-controlled laboratory environment to obtain the accurate perceptual ratings of the CGIs. Then, we propose an effective deep learning-based no-reference (NR) IQA model by utilizing both distortion and aesthetic quality representation. Experimental results show that the proposed method outperforms all other state-of-the-art NR IQA methods on the constructed CGIQA-6k database and other CGIQA-related databases. The database is released at https://github.com/zzc-1998/CGIQA6K.
\end{abstract}




\maketitle

\section{Introduction}
With the rapid development of computer graphics rendering techniques, billions of computer graphics images (CGIs) have been generated and perceived in various applications. Unlike natural scene images (NSIs) that are captured with cameras in the real world, CGIs are rendered through a 2D or 3D model described by means of a computer program \cite{min2017unified,pharr2016physically,zhang2021mesh,zhang2021ano,fan2022no,zhang2022treating}, which has been widely used in architecture, video games, simulators, movies, etc. Different from the generative models based on deep neural networks such as GAN \cite{goodfellow2014generative}, Nerf \cite{mildenhall2020nerf}, etc., computer graphics directly synthesize images using mathematical and computational techniques without learning from the specific training data. According to the rendering techniques and application scenarios, CGIs can be divided into photorealistic images and non-photorealistic images. Photorealistic images are generated to achieve photorealism by modeling the real world \cite{greenberg1997framework}, while non-photorealistic images promote the prosperity of digital art by enabling a wide variety of expressive styles. 
Regardless of the rendering techniques and purposes, both photorealistic and non-photorealistic images inevitably suffer from various distortions, such as texture loss caused by limited computation resources, poor visibility caused by wrong exposure settings, and blur caused by low rendering accuracy, etc. Additionally, a large part of CGIs are transmitted through the network services like cloud gaming and live broadcasting \cite{laghari2019quality}. In most practical cases, CGIs are constantly influenced by compression distortion, which damages the user's Quality of Experience (QoE) \cite{chen2013quality}. Moreover, previous studies \cite{hosu2017konstanz,li2018has,murray2012ava} have shown that the subjective aesthetic preferences of individuals can also influence their quality ratings of images, particularly in relation to CGIs \cite{ling2020subjective}.

\begin{figure}[t]
    \centering
    \subfigure[Sample NSI]{\includegraphics[width = 0.37\linewidth]{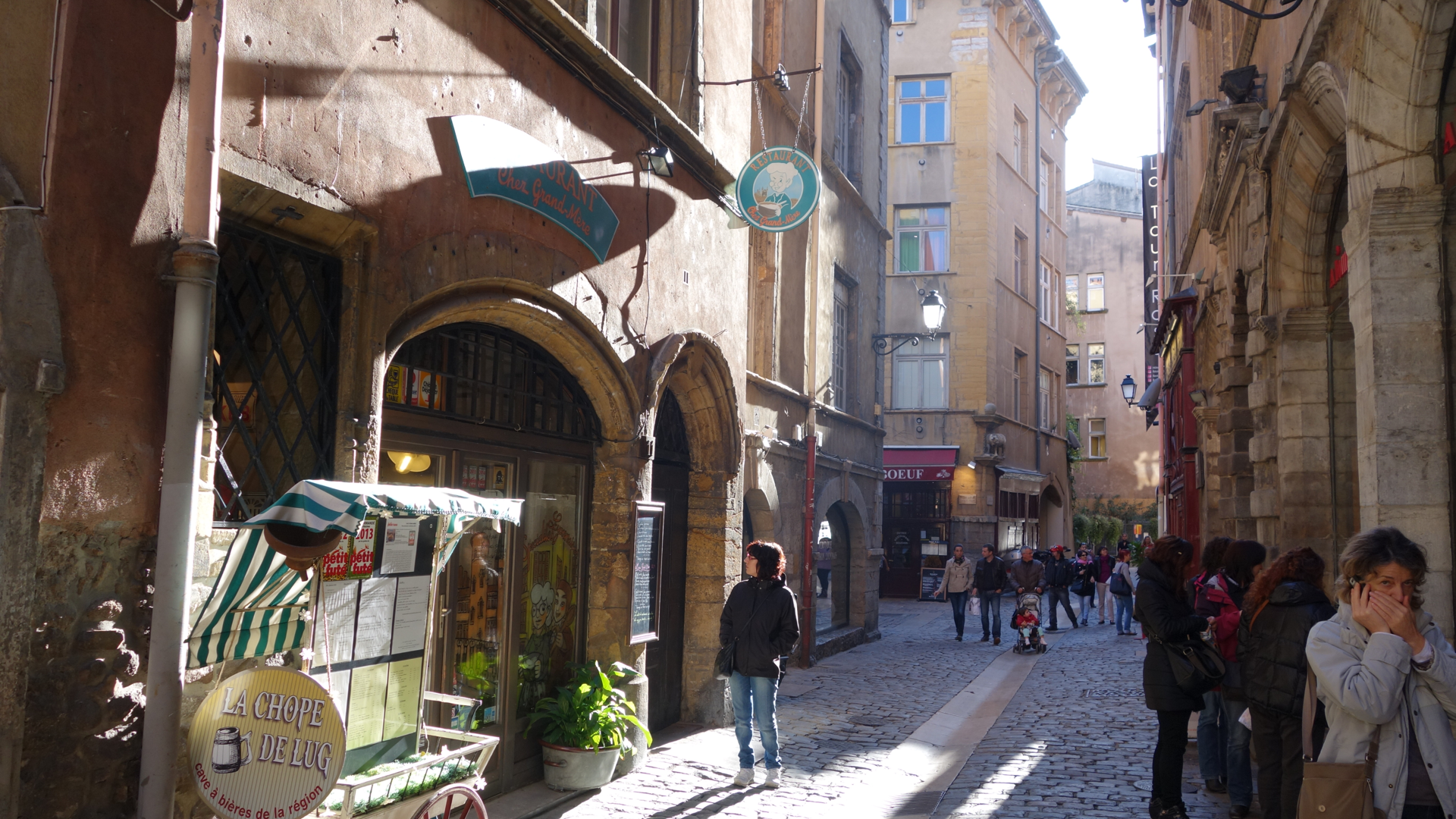}} \hspace{0.5cm}
    \subfigure[Sample CGI]{\includegraphics[width = 0.37\linewidth ]{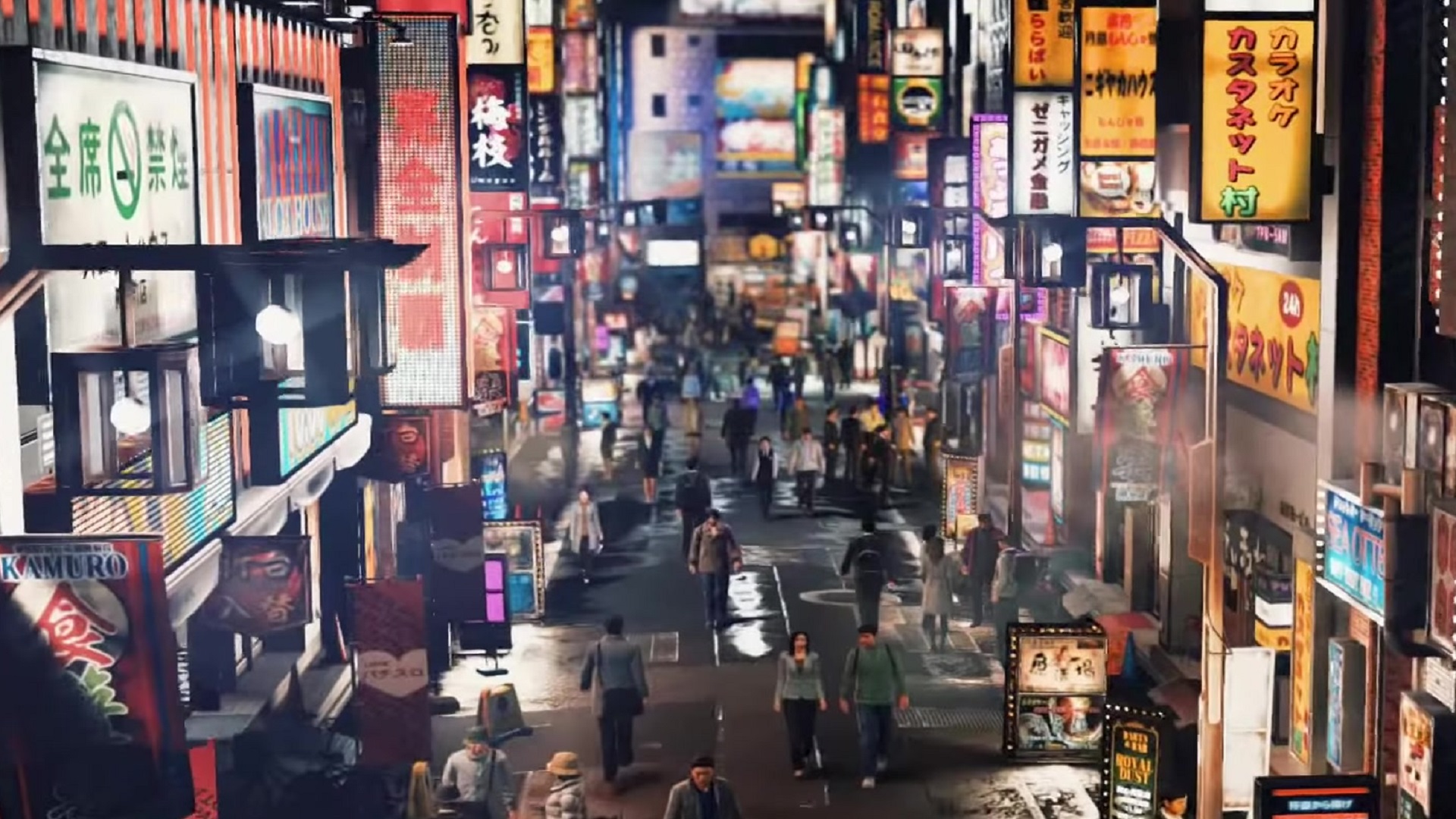}}

    \subfigure[NSI distributions]{\includegraphics[width = 0.4\linewidth]{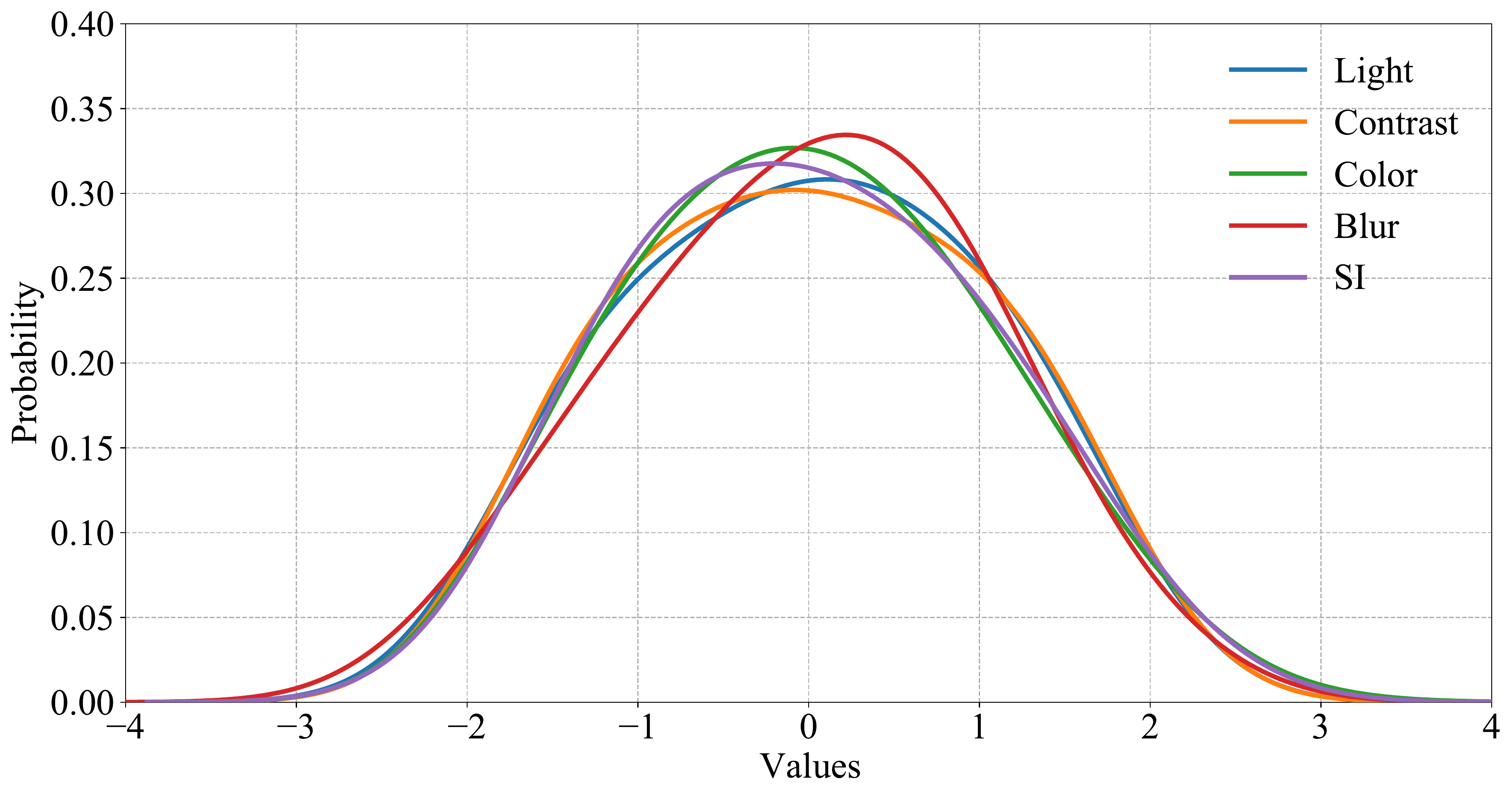}}
    \subfigure[CGI distributions]{\includegraphics[width = 0.4\linewidth]{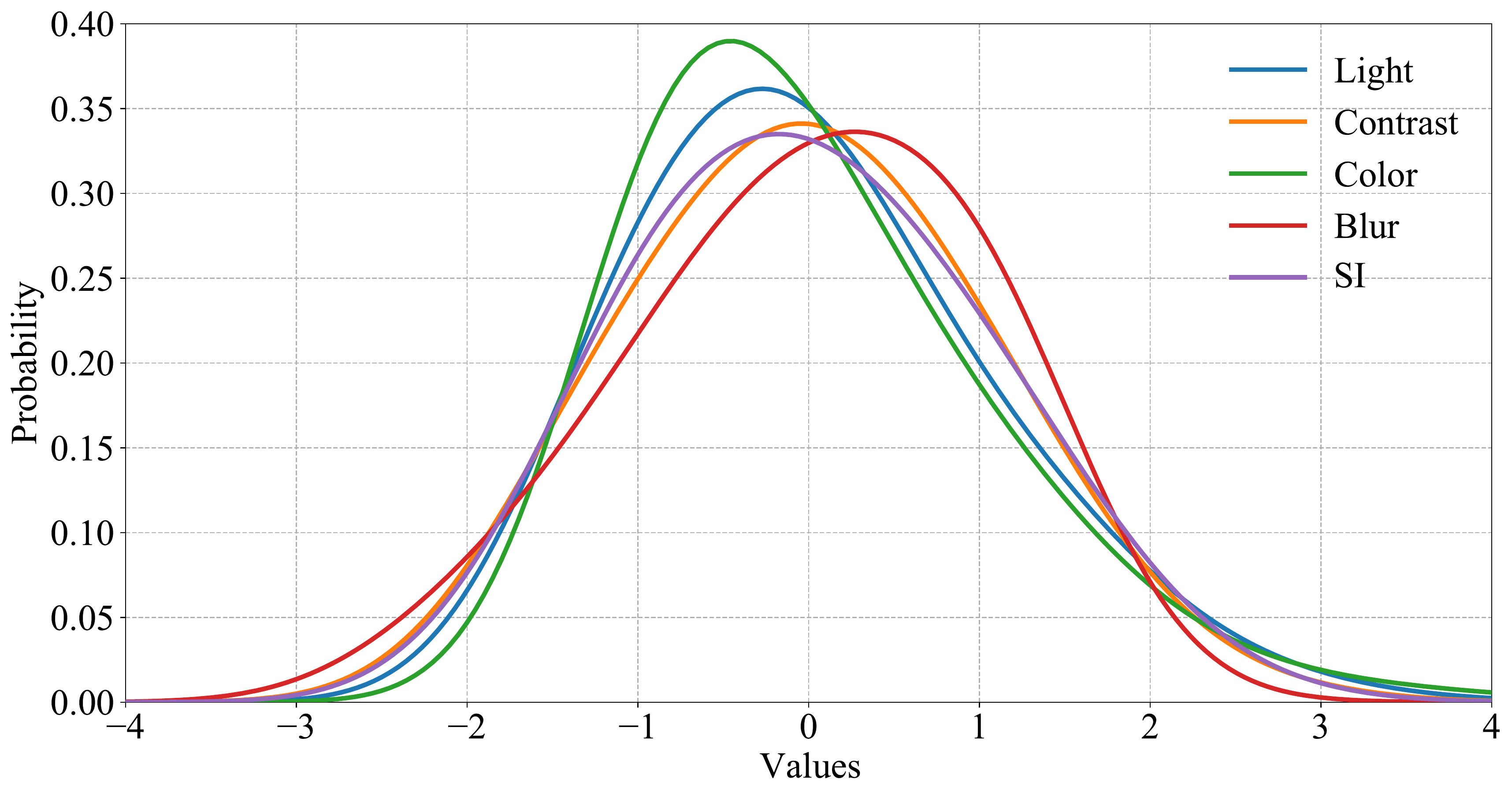}}
    \caption{The normalized probability distributions of the quality-related attributes for NSIs and CGIs. The distributions are obtained from 10,073 NSIs in the KonIQ-10k IQA database \cite{koniq10k} and 18,031 CGIs in the LSCGB database \cite{bai2021robust} respectively. The 'color' indicates the colorfulness of the images and the 'SI' (spatial information) stands for the content diversity of the images.}
    \label{fig:diff}
    \vspace{-0.4cm}
\end{figure}

In the last decade, many image quality assessment (IQA) models have been proposed to tackle quality assessment issues. According to the involving extent of reference images, image quality assessment can be categorized into full-reference (FR), reduced-reference (RR), and no-reference (NR) methods \cite{zhai2020perceptual}. A variety of IQA measures have been proposed \cite{mittal2012no,mittal2012making,li2015no,saad2012blind,min2018blind,narvekar2011no,min2017unified,gu2014using,zhang2018blind,ke2021musiq,su2020blindly,wang2021multi,sun2021blind,peng2021lggd+,song2022blind,jiang2022single,hu2023reduced,zhang2022noreference} and achieved huge success on IQA tasks for natural scene content. 
However, several studies have proven that the state-of-the-art (SOTA) IQA models designed for NSIs are not suitable for images with different statistics distributions from natural scene content \cite{yang2015perceptual,wang2016subjective,ling2020subjective,yu2022subjective}.  Compared with NSIs, CGIs often contain more regular geometric shapes, simpler texture, and relatively less content diversity.  To verify the statistical difference between NSIs and CGIs, we present 5 quality-related attributes distributions of over 10,073 in-the-wild NSIs and  18,031 in-the-wild CGIs for comparison. The NSIs are sourced from the in-the-wild KonIQ-10k IQA database \cite{koniq10k} and the CGIs are collected through the LSCGB database \cite{bai2021robust}. The quality-related attributes include light, contrast, colorfulness, blur, and spatial information (SI), the details description of which can be referred to in \cite{hosu2017konstanz}. Fig. \ref{fig:diff} illustrates the quality-related distributions of the NSIs and CGIs, from which we can see that the normalized probability distributions of NSIs' quality attributes tend to be more centered at zero while the normalized probability distributions of CGIs' quality attributes are relatively irregular in skewness. Specifically, the CGIs contain relatively lower illumination levels and lack colorfulness.

Thus, it is unreasonable to simply transfer IQA models based on NSIs scope to computer graphics IQA (CGIQA). Moreover, the mainstream well-performing IQA methods are mostly designed in the data-driven manner, which highly depends on the quality of corresponding databases. To promote the development of natural scene IQA (NSIQA), various NSI databases have been carried out \cite{sheikh2006statistical,ponomarenko2015image,larson2010most,lin2019kadid,jiang2022underwater}. However, the progress of CGIQA database falls behind: 1) A large part of the publicly available CGIQA-related databases are organized in former times. The selected CGIs are sourced from limited types of media and are generated by outdated rendering techniques, thus such CGIs may not be able to cover the range of current CGIQA tasks. 2) The existing CGIQA-related databases are relatively small in scale, which is difficult for supporting the designing and training of algorithms based on the deep neural networks, which have been the dominant methods for the IQA tasks. 3) Many previous works mainly focus on full-reference (FR) methods and the distortions are manually introduced to the reference images. However, the pristine reference CGIs are usually not available in practical situations and the distortions are complex and unpredictable, which therefore increases the need for evaluating the quality of in-the-wild CGIs in the NR manner.

Therefore, to address the above challenges, we first establish a large-scale CGIQA database containing 6,000 CGIs.  In order to construct the benchmark for the CGIQA task, we select several SOTA NR IQA models including handcrafted-based and deep learning-based methods for comparison. What's more, we propose a deep learning-based NR IQA model through both distortion and aesthetic aspects to help promote the performance on the proposed CGIQA database. The experimental results show that the proposed method is more effective for predicting the quality scores of CGIs. The major contributions of our work are summarized as follows:
\begin{itemize}
    \item To the best of our knowledge, we construct the largest in-the-wild CGIQA database (CGIQA-6k), which consists of 6,000 CGIs from games and movies along with the subjective quality scores. Our database covers a wider range of resolutions (480p$\sim$4K) and contains more diverse content as well as authentic distortions. 
    \item We especially design a deep learning-based model by jointly employing distortion and aesthetic quality representation. The distortion stream integrates multi-stage feature fusion and multi-stage channel attention mechanisms, aimed at improving the learning of distortion representation. Meanwhile, the aesthetic stream leverages prior knowledge acquired from the aesthetic quality assessment (AQA) databases. The integration of these two streams empowers the model with an enhanced understanding of the various quality levels exhibited by CGIs. 
    \item The proposed method and some mainstream NR IQA models are validated on the CGIQA-6k database along with other CGIQA-related databases to provide the benchmark for CGIQA tasks. The ablation study, cross-database validation, and statistical test are conducted for further analysis. In-depth discussions about the experimental performance are given as well.
\end{itemize}

The remainder of this paper is organized as follows. Section \ref{sec:related} gives a brief introduction of the related work. Section \ref{sec:subjective} describes the establishment of the CGIQA-6k database and the subjective experiment. Section \ref{sec:method} presents the details of the proposed IQA model. Section \ref{sec:experiment} gives the experimental performance of the proposed method and other SOTA NR IQA models. The ablation study and statistical test are also conducted and discussed. Section \ref{sec:conclusion} concludes this paper.

\section{Related Work}
\label{sec:related}
In the section, we briefly summarize the development of NR IQA models as well as the construction of CGIQA-related databases.

\subsection{No-reference Image Quality Assessment}
\vspace{0.1cm}
Generally speaking, the NR IQA models can be categorized into handcrafted-based methods (extracting features in handcrafted manners) and deep learning-based methods (extracting features using deep neural networks), both of which have been confirmed to be effective for common IQA tasks.
BRISQUE \cite{mittal2012no} employs natural scene statistics (NSS) in the spatial field for quality analysis. NIQE \cite{mittal2012making} makes use of measurable deviations from statistical regularities observed in natural images for evaluation. BLIINDS2 \cite{saad2012blind} further uses NSS approach in the discrete cosine transform (DCT) domain for quality assessment.  CPBD \cite{narvekar2011no} calculates the blur levels by computing the cumulative probability of blur detection. BIBLE \cite{li2015no} focuses on blur-specific distortions based on discrete orthogonal moments. NFERM \cite{gu2014using} studies the quality of images by using free energy principle. UCA \cite{min2017unified}  is a unified content-type adaptive blind IQA measure aiming at compression distortion.

With the development of deep neural networks, many deep learning-based IQA methods have been proposed. DBCNN \cite{zhang2018blind} constitutes two streams of deep neuron networks and deals with both synthetic and authentic distortions.  HyperIQA \cite{su2020blindly} uses a self-adaptive hyper network to deal with the challenges of distortion diversity and content variation for IQA issues. MUSIQ \cite{ke2021musiq} employs a multi-scale image quality transformer to represent image quality levels at different granularities. MGQA \cite{wang2021multi} and StairIQA \cite{sun2021blind} both hierarchically integrate the features extracted from intermediate layers to take advantage of both low-level and high-level visual information. Most of the mentioned IQA measures mentioned above have shown strong ability of predicting quality levels for NSIs on the traditional IQA databases such as LIVE \cite{sheikh2006statistical}, TID2013 \cite{ponomarenko2015image}, CSIQ \cite{larson2010most}, Kadid10K \cite{lin2019kadid}, etc.

\subsection{No-reference Aesthetic Quality Assessment}
{
Earlier methods in AQA use handcrafted feature extraction. For instance, they utilize features like color arrangement, explosion, sharpness, visual saliency, composition, and focusing, as mentioned by \cite{datta2006}, \cite{ke2006}, \cite{nishiyama2011}, \cite{sun2009}, and \cite{zhang2014}.

Nowadays, the trend is to employ deep learning models for aesthetic evaluation. Studies like \cite{lu2015}, \cite{talebi2018}, and \cite{yang2019} take smaller image patches from resized images as input for further feature extraction. However, a challenge with this approach is that randomly chosen patches, even when combined, don't capture both the local and the overall image details.
Addressing this, \cite{ma2017} creates an innovative model that uses an object-based attribute graph to better understand the image layout. \cite{zhang2019} develops an attention mechanism to zoom in on specific aesthetic-related areas. \cite{yang2019} uses a network to prioritize local views based on their saliency. \cite{sheng2018} devises an attention module that gauges the importance of each image patch.
A notable approach by \cite{hou2020} incorporates region-of-interest (RoI) features from a generic object detector, leading to a more effective blend of overall and localized features. Another study \cite{cui2018} posits that a single label might not sufficiently assess the aesthetic quality of an image, and hence suggests using a range of quality levels. \cite{shu2021} presents a method that uses partial attribute annotations to address the manual attribute annotation shortage. \cite{kong2016} focuses on Knowledge Distillation, enabling the AQA model to discern and extract more relevant aesthetic patterns.}

{Both IQA and AQA aim to assess an image's merits but differ in focus and application. IQA concentrates on technical attributes like sharpness and noise which is vital in fields requiring image clarity, like camera imaging, medical imaging, and image compression/transmission systems. Feedback in IQA is technical, suggesting adjustments in equipment settings or resolution. On the other hand, AQA delves into an image's artistic attributes like composition and emotional impact, central in fields like photography or art. Feedback in AQA is more abstract, recommending changes in color scheme or subject matter. While both evaluations have objective measures, they inherently contain subjective elements, influenced by individual perceptions.}

\begin{table*}[t]\footnotesize
\renewcommand\tabcolsep{2pt}
\centering
\caption{The comparison of previous CGIQA-related databases and our database. The scale indicates the number of stimuli with mean opinion scores (MOSs).}
\vspace{-0.2cm}
\begin{tabular}{ccccccc}
\toprule
Database    &Year   & Scale & Source                & Scope             & Resolution Range   &Public  \\
\midrule
CCT \cite{min2017unified}    &2017       & 528   & PC game images        & Compression distortion       & 720P$\sim$1080P    &Yes       \\
GamingVideoSET \cite{barman2018gamingvideoset}&2018 & 90 & PC game videos & Compression distortion  &480P$\sim$1080P &Yes\\
KUGVD \cite{barman2019no} &2019 & 90    & PC game videos      & Downsampling \& Bitrates control  &480P$\sim$1080P   & Yes    \\
CGVDS \cite{zadtootaghaj2020quality} & 2020 & 225 & PC game videos & Compression distortion  &480P$\sim$1080P   & Yes\\
TGQA \cite{ling2020subjective}   &2020       & 1091  & Mobile game images    & Aesthetic evaluation         & 1080P   &Yes               \\

TGV \cite{wen2021subjective}    &2021       & 1293  & Mobile game videos    & Stull \& Bitrates control              & 480P$\sim$1080P  & No\\
LIVE-YT-Gaming \cite{yu2022subjective} &2021 & 600 & PC game videos   & UGC distortions & 360P$\sim$1080P   &Yes \\
NBU-CIQAD \cite{chen2021perceptual} & 2021 & 2700 & Cartoon images & Brightness, Saturation, Contrast & 1080P & Yes\\
CGIQA-6k(ours)     &2023      & 6000  & Games \& Movies & in-the-wild distortion      & 480P$\sim$4K  &Yes \\
\bottomrule
\end{tabular}

\label{tab:compare}
\end{table*}

\begin{figure*}[t]
    \centering
    \includegraphics[width = \linewidth]{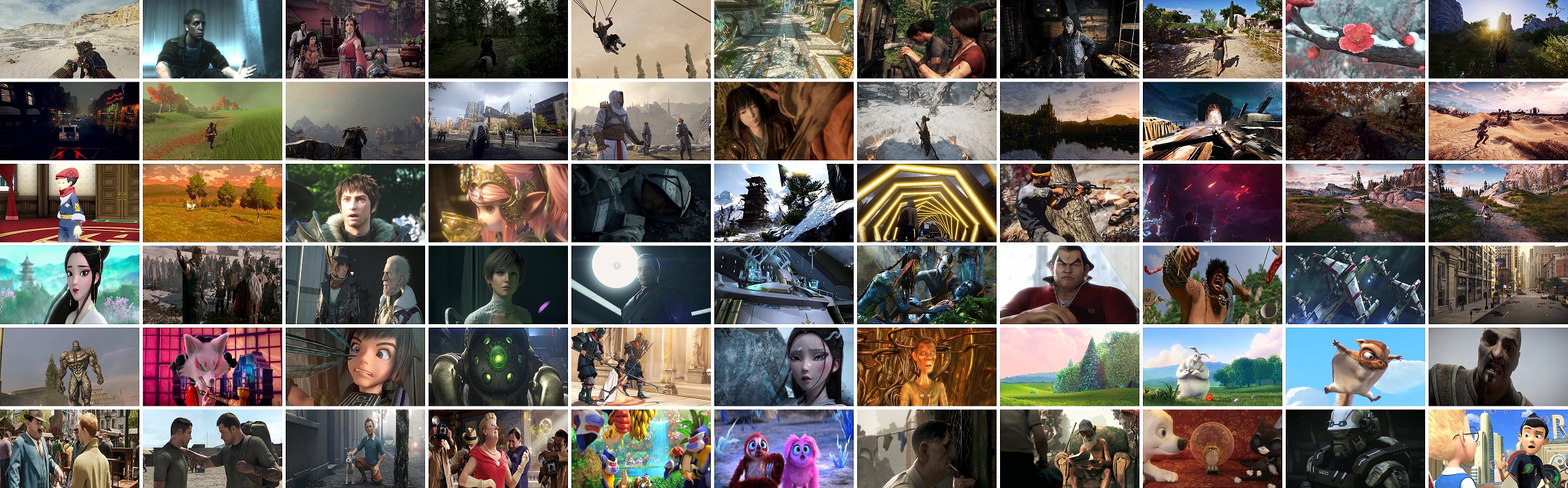}
    \caption{Sample images from the CGIQA-6k database. Top three rows: CGIs from games. Bottom three rows: CGIs from movies.}
    \label{fig:overview}
\end{figure*}

\subsection{CGIQA-Related Databases}
The CCT database \cite{min2017unified} is a cross-content-type database including natural scene images, screen content images, and computer graphics images, which mainly focuses on compression distortion. The GamingVideoSET database \cite{barman2018gamingvideoset} contains 576 compressed gaming videos from 24 raw gaming videos and conducts the subjective quality assessment experiment on 90 compressed gaming videos. Similarly, the KUGVD database \cite{barman2019no} obtains 144 distorted gaming videos from 6 reference gaming videos and provides subjective ratings for 90 distorted gaming videos. The CGVDS database \cite{zadtootaghaj2020quality} is carried out for cloud gaming applications and provides 225 gaming videos along with quality scores. The TGQA database \cite{ling2020subjective} focuses on the aesthetic assessment of mobile game images and presents the subjective study for 1,091 mobile game images on multi-dimensional aesthetic factors. The TGV database \cite{wen2021subjective} generates 1,293 mobile gaming sequences encoded with three codecs along with quality scores. The LIVE-YT-Gaming database \cite{yu2022subjective} pays more attention to the user-generated content (UGC) gaming videos and consists of 600 authentic UGC gaming videos with subjective ratings. The NBU-CIQAD database \cite{chen2021perceptual} deals with the quality assessment of cartoon images and contains 2,600 distorted cartoon images with quality scores.
A detailed comparison of the CGIQA-related quality assessment database is given in Table \ref{tab:compare}. Through further observation, it can be clearly found that our database is the largest in scale and covers the most common resolutions.


\section{Subjective Quality Assessment for Computer Graphics Images}
\label{sec:subjective}
Few large-scale databases for CGIs have been developed in the quality assessment field. To further facilitate research, we construct a large-scale database called CGIQA-6k and conduct the corresponding subjective experiment to derive the subjective quality scores. 



\subsection{Data Collection}
The previously published LSCGB database for CGI and NSI detection \cite{bai2021robust} contained 18,031 CGIs sourced from popular 3D games and 3D movies. To cover more range of resolutions, we personally collect 40,000 CGIs through the frames of stream videos from YouTube, Netflix, and Bilibili with different playback settings. Furthermore, we obtain about 10,000 CGIs by recording screenshots of local game demos. To reduce the effect of uncorrelated distractions, we specifically eliminate the life bars, mini maps, and subtitles. To ensure the diversity of content, we also manually remove the CGIs with close content. Considering different viewpoints can provide different experiences even in interactive games, we include both first-person view (POV) CGIs and third-person view (TOV) CGIs. The POV CGIs allow the viewers to perceive the scene through the character's eyes, providing the most immersive feelings \cite{denisova2015first}.  The TOV CGIs give a broader view of the environment and enable the viewers to have a clear sight of the main character \cite{voorhees2012guns}. These mentioned two views make up the majority views of mainstream video games on the market. As the CGIs focus more on storylines rather than interactivity, we choose mostly portrait CGIs and landscape CGIs for evaluation, which are commonly found in movies and game cutscenes. 

\begin{figure*}[t]
    \centering
    \subfigure[Game]{\includegraphics[width=.3\linewidth]{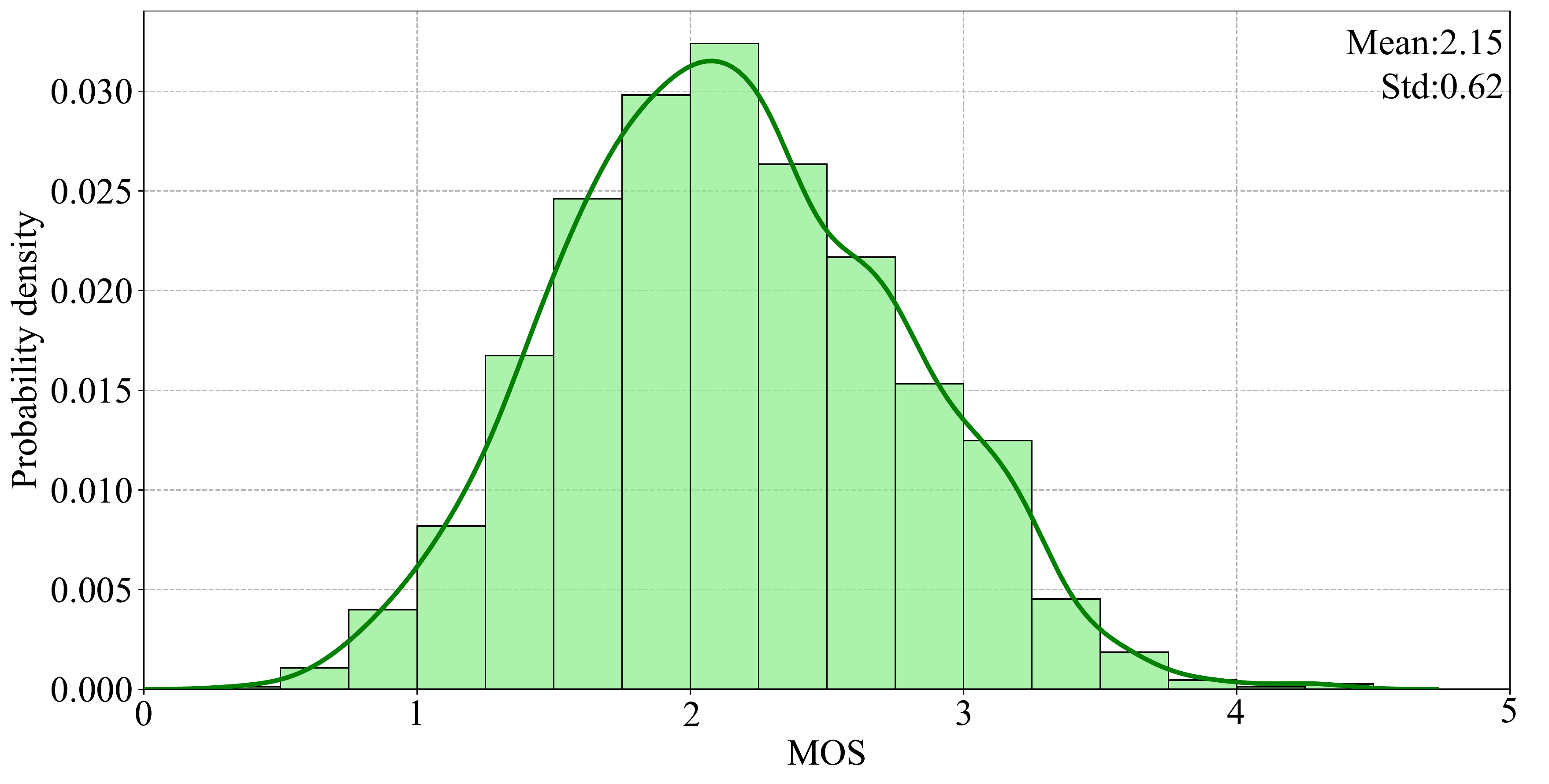}}
    \subfigure[Movie]{\includegraphics[width=.3\linewidth]{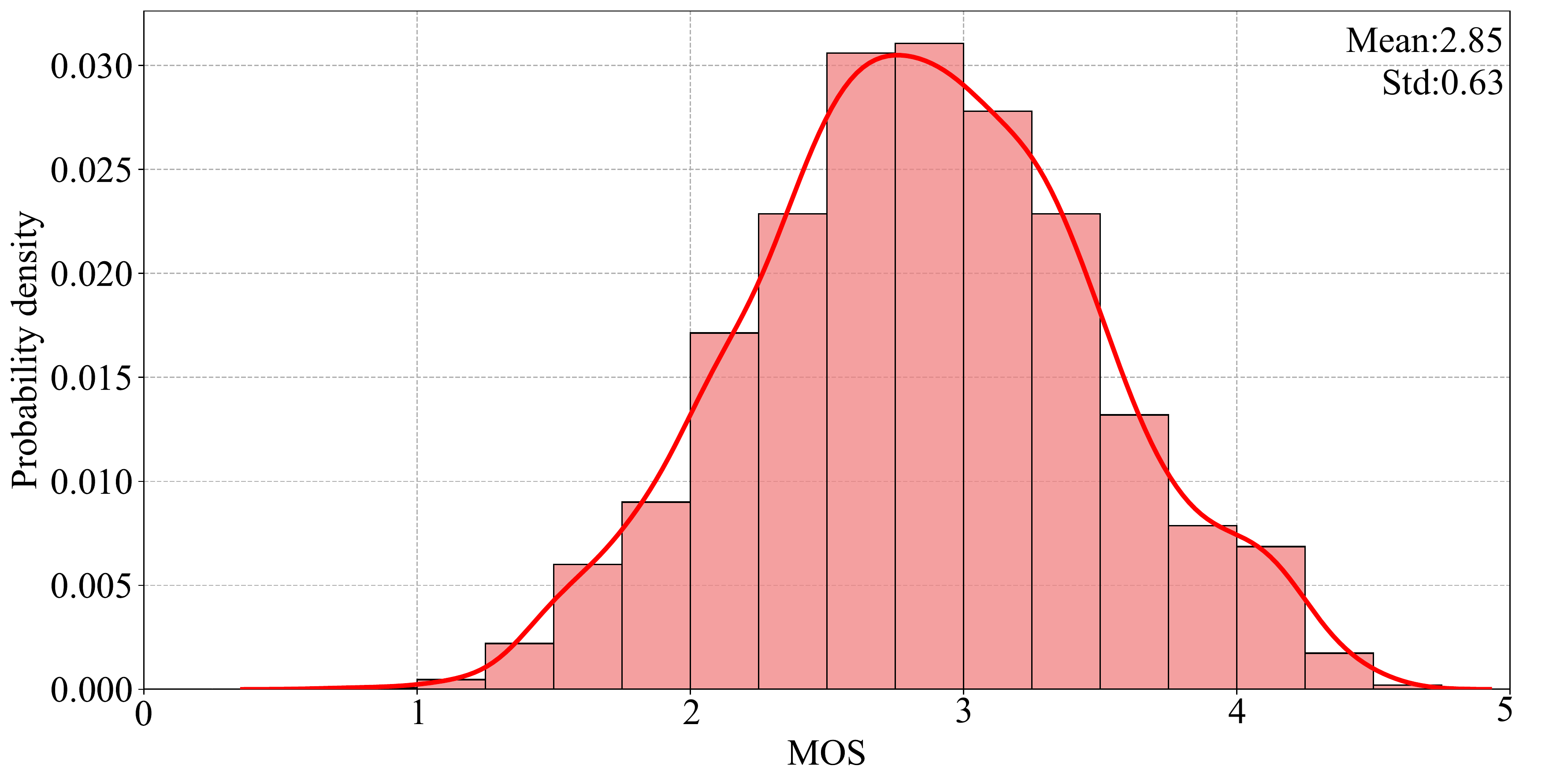}}
    \subfigure[All]{\includegraphics[width=.3\linewidth]{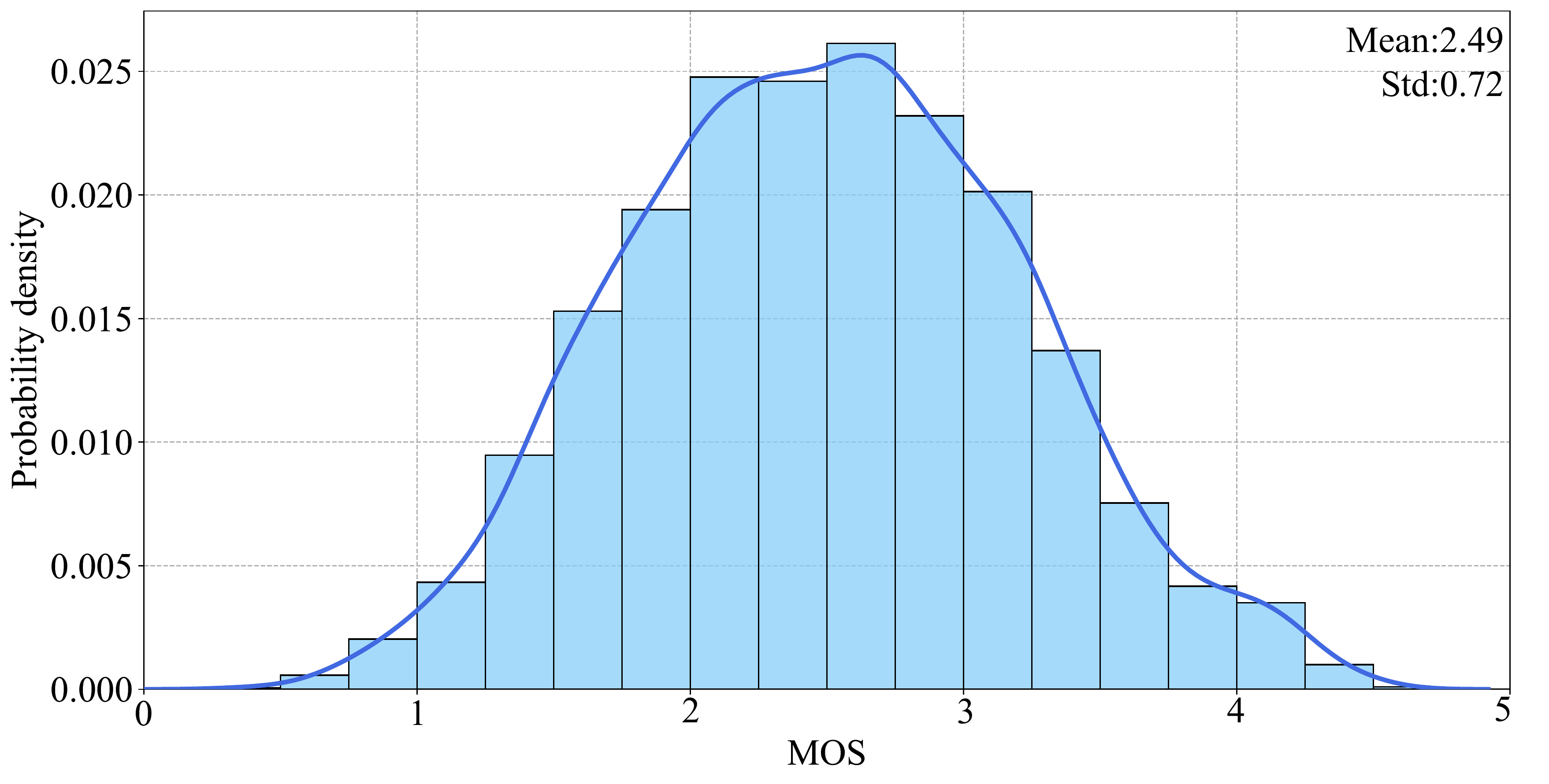}}
    \caption{The distribution of the MOSs of the CGIQA-6k database. (a) represents the MOS distribution for game CGIs while (b) indicates the MOS distribution for movie CGIs. (c) exhibits the MOS distribution for all CGIs. Additionally, the mean and standard deviation values of the MOSs are marked on the top right as well.}
    \label{fig:mos}
    \vspace{-0.4cm}
\end{figure*}

\begin{figure}[t]
    \centering
    \subfigure[Confusing exposure, MOS = 1.14]{\includegraphics[width=0.4\linewidth]{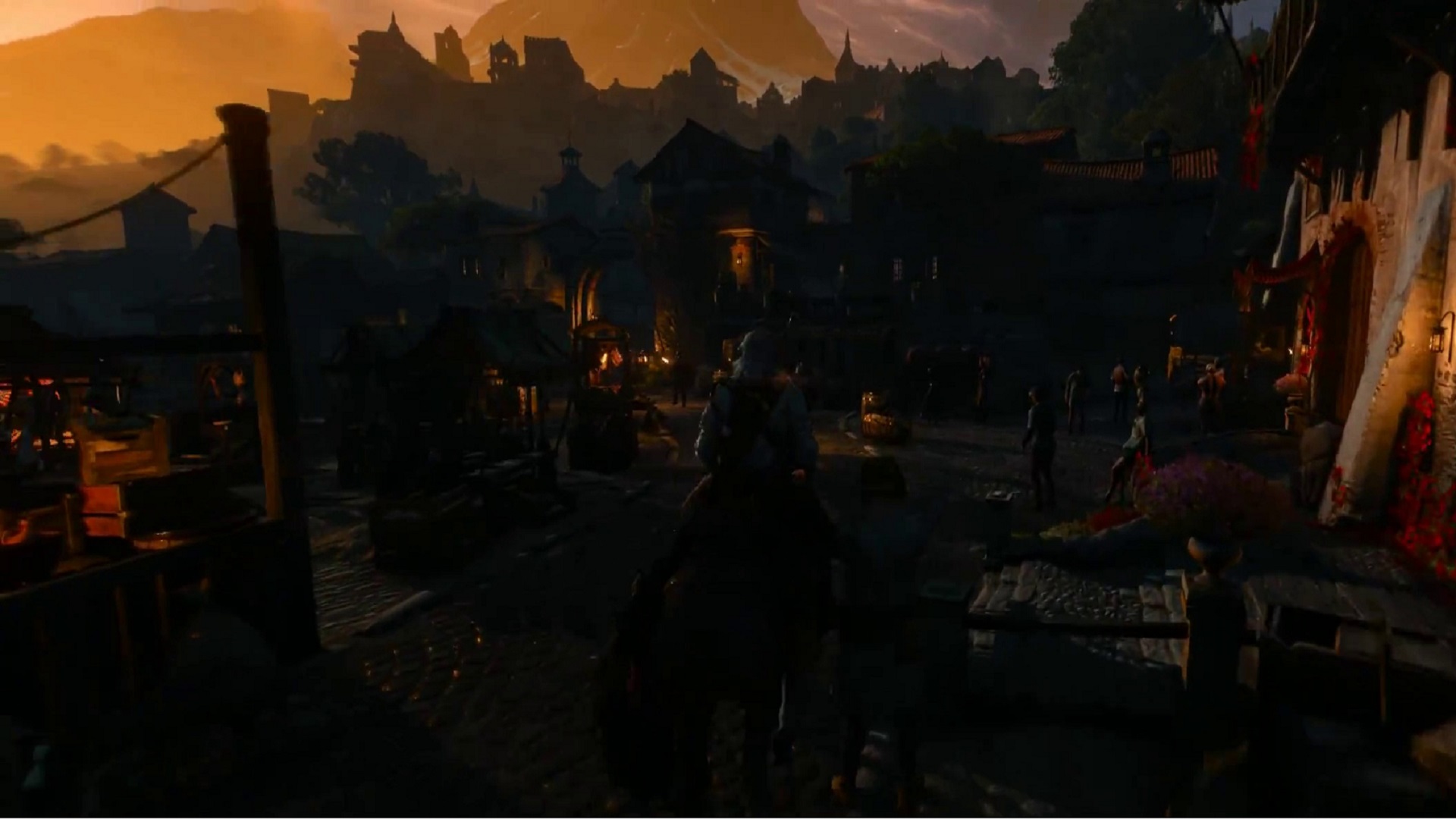}\label{fig:distortions-a}}
    \subfigure[Texture loss, MOS = 1.50]{\includegraphics[width=0.4\linewidth]{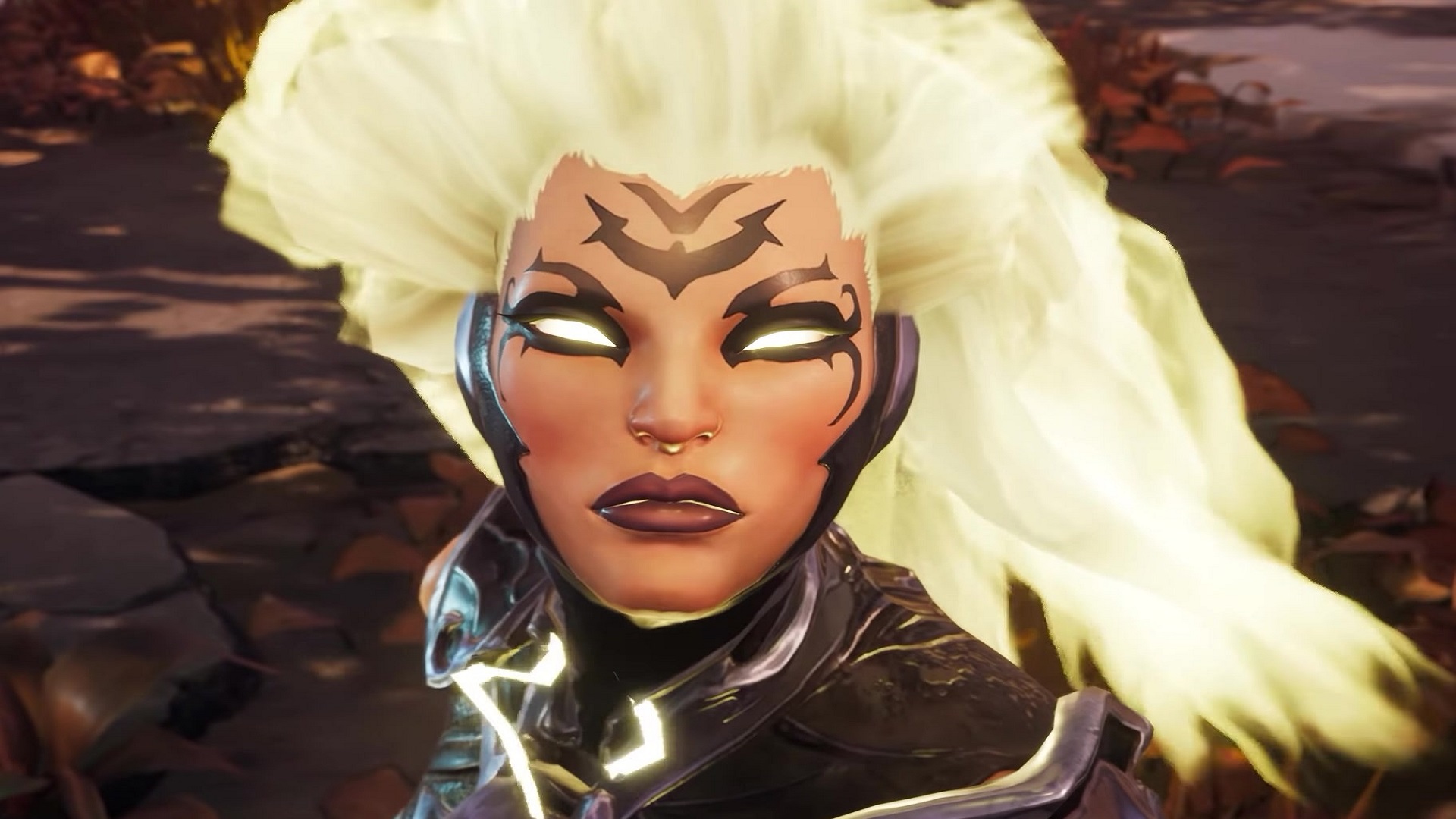}\label{fig:distortions-b}}
    \subfigure[Low rendering accuracy, MOS = 1.29]{\includegraphics[width=0.4\linewidth]{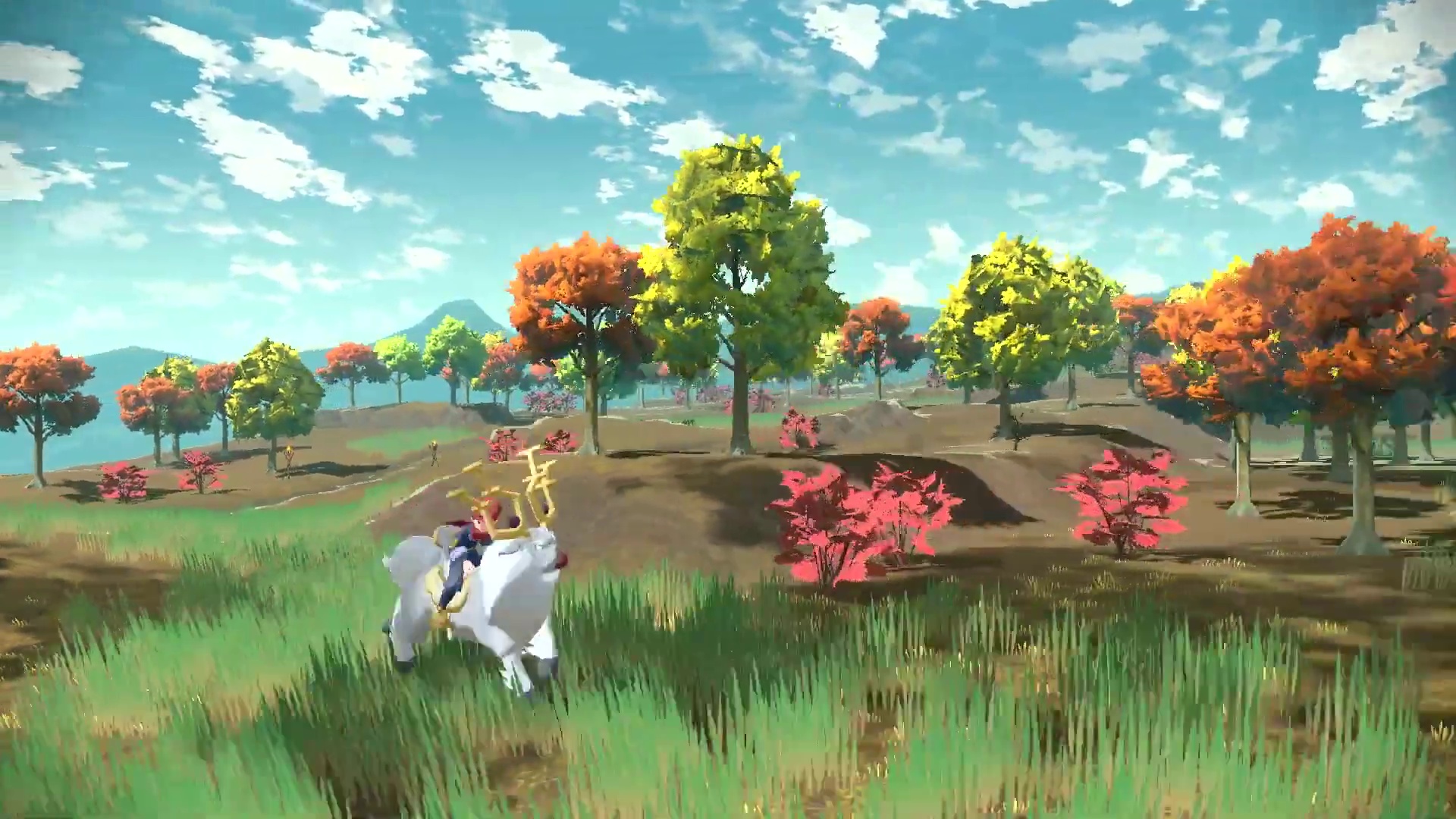}\label{fig:distortions-c}}
    \subfigure[Compression distortion, MOS = 1.51]{\includegraphics[width=0.4\linewidth]{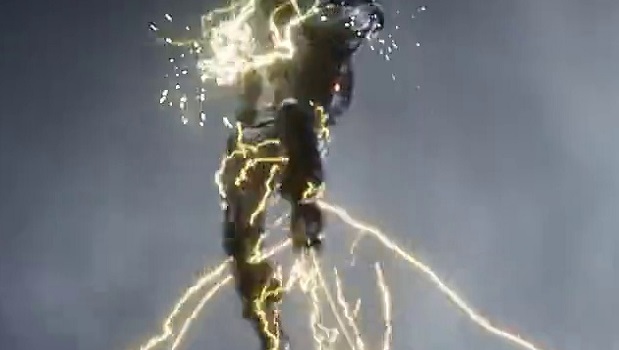}\label{fig:distortions-d}}
    \caption{Examples of some typical distortions of CGIs. The content of (a) is poorly visible due to the confusing exposure and the hair texture is quite ambiguous in (b). The components of (c) exhibit relatively simple geometry shapes with jagged edges caused by low rendering accuracy and the compression distortion damages the perceptual quality of (d). The subjective quality scores for (a), (b), (c), and (d) are provided and the MOS ranges from 0-5. }
    \label{fig:distortions}
\end{figure}

\subsection{Data Sampling}
\begin{figure}[!t]
\centering
\subfigure[Higher aesthetic quality and more ambiguous, MOS = 2.71]{\includegraphics[width=0.4\linewidth,height=0.23\linewidth]{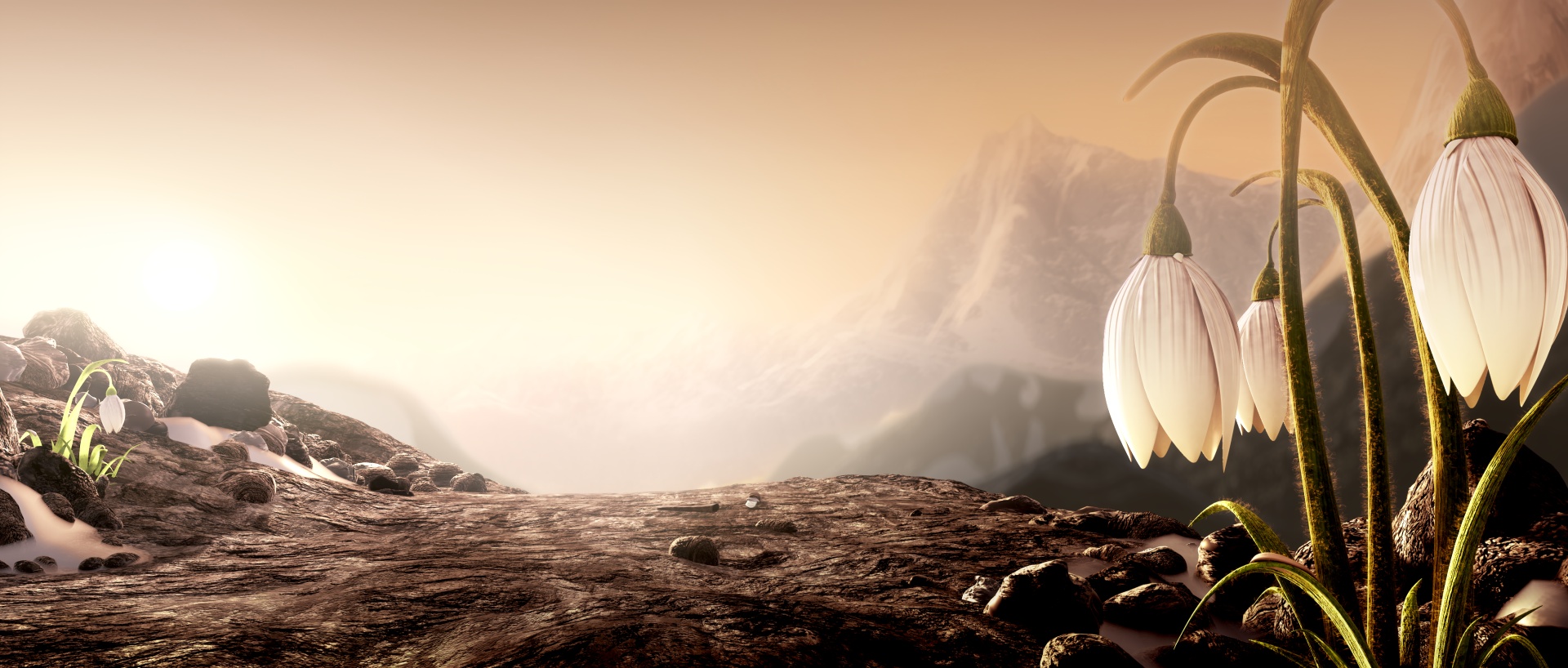}%
} 
\subfigure[Lower aesthetic quality and clearer content, M0S = 2.45]{\includegraphics[width=0.4\linewidth,height=0.23\linewidth]{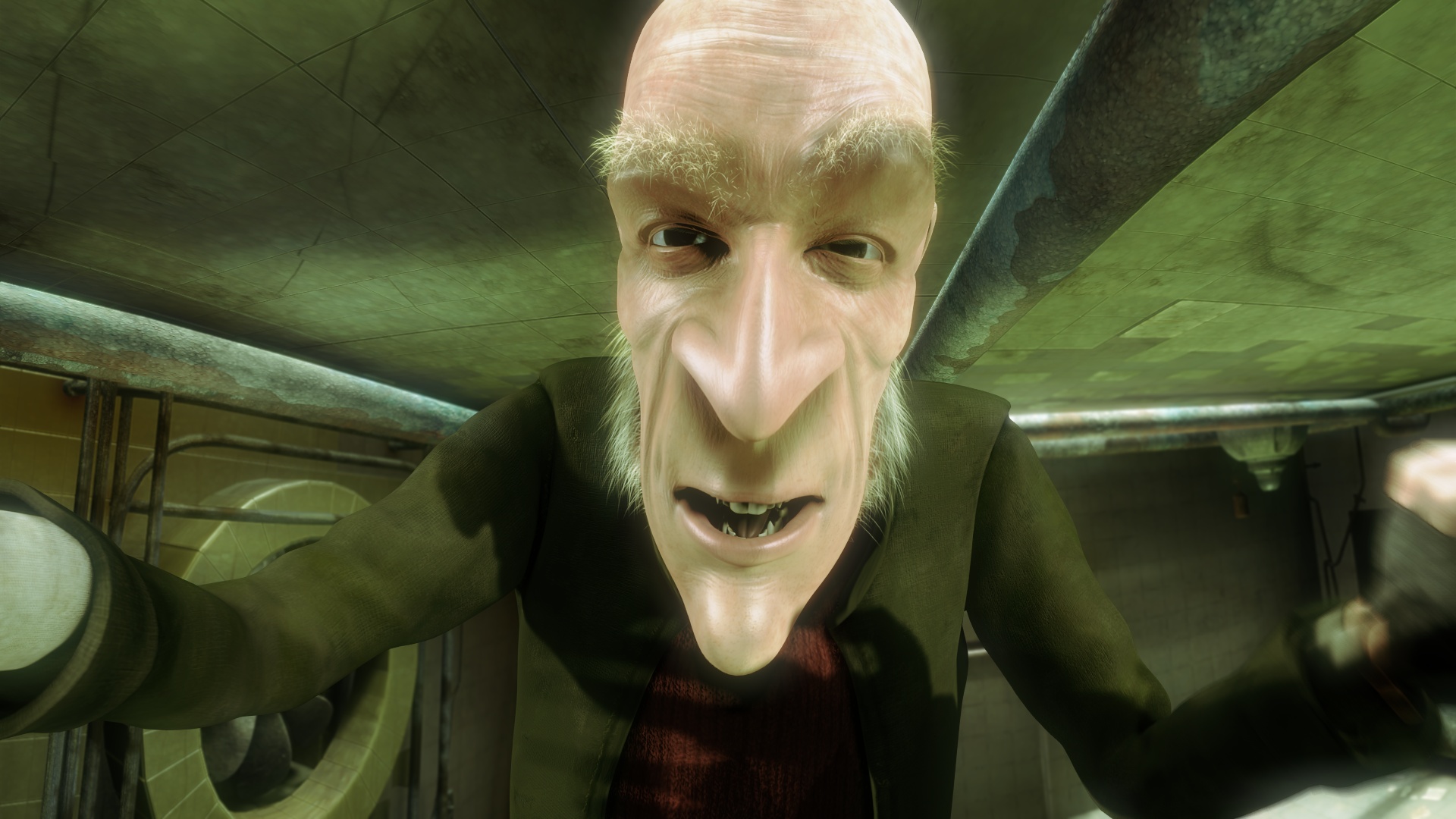} 
}

\caption{Examples of CGIs with higher and lower aesthetic quality. Although the content of (a) appears more vague and ambiguous while the details of (b) appear clearer, (a) elicits a higher quality experience than (b) from the aesthetic perspective in terms of MOS.}
\label{fig:aesthetic}
\vspace{-0.5cm}
\end{figure}

After conducting the aforementioned data collection process, we successfully acquired approximately 55,000 CGIs. To ensure the preservation of the quality distribution, we follow the recommended approach presented in previous works \cite{wu2023exploring, hosu2017konstanz, Ying2021, zhang2023md}. Accordingly, we proceed to randomly sample a subset of CGIs, while ensuring that the distributions of quality-related attributes (including light, contrast, colorfulness, blur, and spatial information \cite{hosu2017konstanz}) within the subset align with those observed in the original collected CGIs. Finally, we obtain a total of 6,000 CGIs, which consists of 3,000 game CGIs and 3,000 movie CGIs. Samples of the selected CGIs are exhibited in Fig. \ref{fig:overview}, where the top three rows are game CGIs and the bottom three rows are movie CGIs.
Generally speaking, most CGIs have been processed through compression or transmission systems, we do not introduce new distortions in addition and pay most attention to the in-the-wild distortions of CGIs. {Examples of typical distortions are shown in Fig. \ref{fig:distortions}: In CGI (a), the content is poorly visible due to inappropriate exposure. Such issues with exposure could stem from incorrect settings in the lighting model or improper simulation of reflection and scattering effects during rendering. In CGI (b), the hair texture appears ambiguous. Rendering hair is challenging in computer graphics because hair consists of countless tiny fibers, each interacting with light.
Components in CGI (c) present relatively simplistic geometric shapes but suffer from jagged edges, a consequence of low rendering accuracy. Furthermore, the perceptual quality of the CGI is compromised by compression distortion. In computer graphics, CGIs are often compressed to save on storage space or transmission bandwidth. However, if the compression rate is too high or if an unsuitable compression algorithm is used, noticeable distortions can be introduced, degrading the perceived quality of CGIs.} Examples of different aesthetic quality levels are shown in Fig. \ref{fig:aesthetic}.

\subsection{Subjective Experiment Methodology}
\begin{figure}[t]
    \centering
    \includegraphics[width=0.6\linewidth]{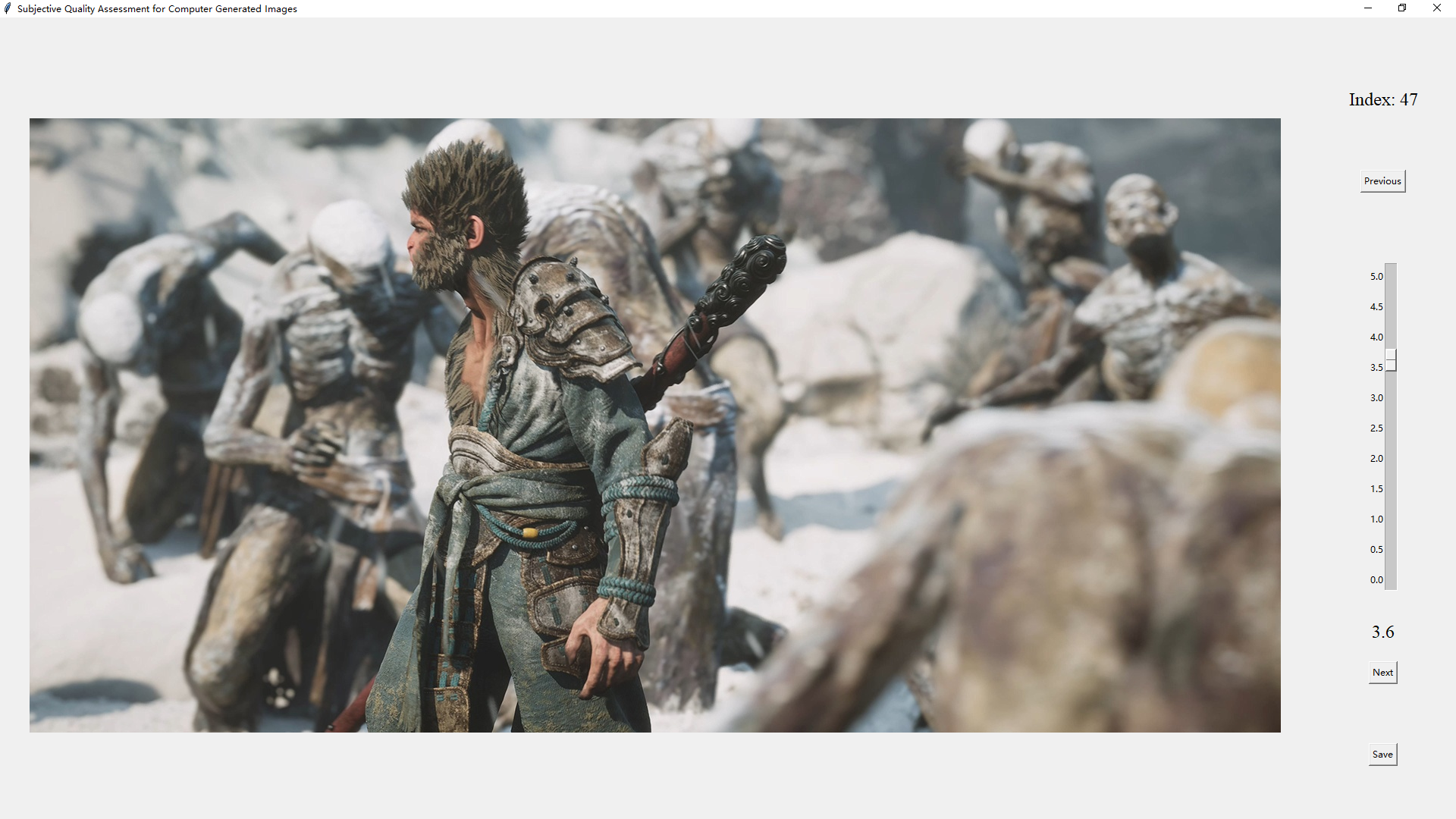}
    \caption{Screenshot of the interface for subjective experiments.}
    \label{fig:interface}
    \vspace{-0.5cm}
\end{figure}

To obtain the perceptual quality scores of CGIs, the subjective experiment is conducted with the recommendations of ITU-R BT.500-13 \cite{bt2002methodology}. All CGIs are shown in random order with an interface designed by Python Tkinter on an iMac monitor which supports a resolution up to 4096 $\times$ 2304. The screenshot of the interface is illustrated in Fig. \ref{fig:interface}, from which we can see that the interface allows users to freely browse the previous and next CGIs, and record the quality scores through a scroll bar. A total of 60 graduate students (38 males and 22 females) are invited to participate in the subjective experiment. The viewers are seated at a distance of around 1.5 times the screen height (45cm) in a laboratory environment with normal indoor illumination. The quality scale scores from 0 to 5, with a minimum interval of 0.1.

The participants are instructed to assess the CGIs based on both distortion and aesthetic quality, providing ratings with an overall quality score.
The whole experiment is split into 20 sessions, each of which includes the subjective quality evaluation for 300 CGIs. In this way, we limit the experiment time for each session to less than half an hour. Every session is attended by at least 20 viewers so that every CGI is evaluated by at least 20 subjects, which generates more than 20$\times$6,000=120,000 quality ratings in total.

\subsection{Subjective Data Processing}
After the subjective experiment, we collect all the quality ratings from the subjects. Let $r_{ij}$ denote the raw rating judged by the $i$-th subject on the $j$-th image, the z-scores are obtained from the raw ratings as follows:
\begin{equation}
z_{i j}=\frac{r_{i j}-\mu_{i}}{\sigma_{i}},
\end{equation}
where $\mu_{i}=\frac{1}{N_{i}} \sum_{j=1}^{N_{i}} r_{i j}$, $\sigma_{i}=\sqrt{\frac{1}{N_{i}-1} \sum_{j=1}^{N_{i}}\left(r_{i j}-\mu_{i}\right)}$, and $N_i$ is the number of images judged by subject $i$.
Then we remove the ratings from unreliable subjects by employing the recommended subject rejection procedure described in the ITU-R BT.500-13 \cite{bt2002methodology}.
The mean opinion score (MOS) of the image $j$ is computed by averaging the rescaled z-scores: 
\begin{equation}
M O S_{j}=\frac{1}{M} \sum_{i=1}^{M} z_{i j}^{'},
\end{equation}
where $M O S_{j}$ indicates the MOS for the $j$-th CGI, $M$ is the number of the valid subjects, and $z_{i j}^{'}$ are the rescaled z-scores. The corresponding MOS distributions are exhibited in Fig. \ref{fig:mos}. It can be evidently seen that the MOS distributions follow Gaussian-like distribution. With closer inspections, the movie CGIs tend to gain higher quality scores than the game CGIs, which is consistent with the practical situations that games are usually rendered in real-time manners while movies often have more resources for rendering.

\begin{figure*}
    \centering
    \includegraphics[width = \linewidth]{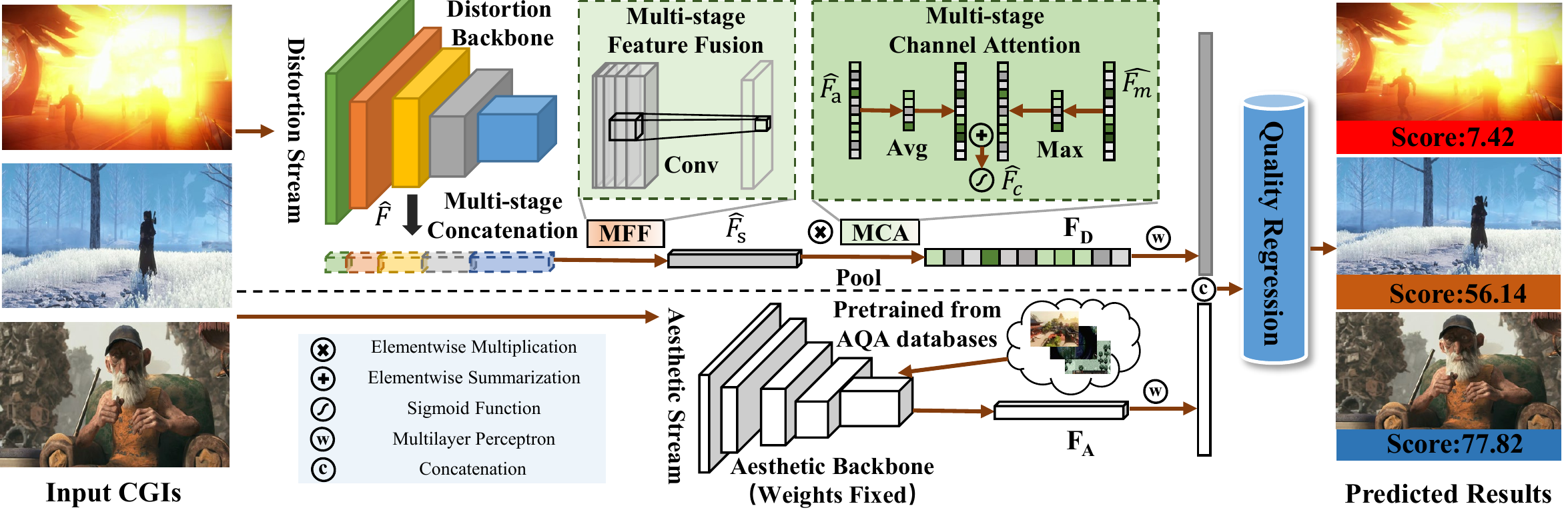}
    \caption{The framework of the proposed method, where the proposed MFF and MCA modules are arranged in a series manner and the aesthetic stream backbone's weights are fixed during the training. }
     \label{fig:framework}
     \vspace{-0.3cm}
\end{figure*}

\section{Proposed Method}
\label{sec:method}
In this paper, we propose an effective deep learning-based method to evaluate the quality of CGIs, the framework of which is clearly shown in Fig. \ref{fig:framework}. The framework consists of two streams: the distortion stream and the aesthetic stream. The distortion stream utilizes the multi-stage feature fusion (MFF) module and the multi-stage channel attention (MCA) module to enhance the learning of broad-range distortion representation, which has been proven effective in many IQA tasks \cite{golestaneh2020no,you2022explore,liu2022quality,hu2020subjective,lu2022deep,wang2022deep}. The aesthetic stream employs the vision backbones pre-trained on the mobile game aesthetic quality assessment TGQA \cite{ling2020subjective} and general aesthetic quality assessment AVA \cite{murray2012ava} databases to extract aesthetic quality-aware features. Finally, the distortion and aesthetic features are fused and regressed into quality scores. 

\subsection{Distortion Stream}
\subsubsection{Multi-stage Feature Fusion}
\label{sec:MFF}
Games and movies are among the most critical artistic mediums since they are believed to pass conceptual ideas, spread emotional powers, and inspire creative thinking \cite{gintere2019new,patton2013games,mackay2017fantasy,mendiburu20123d}. Therefore, the CGIs obtained from games and movies require the viewers to identify the semantic contents before fully understanding the images. In previous quality assessment studies \cite{dhar2011high,kao2017deep,zhang2023mm}, the semantic information is stated to be closely correlated with the high-level attributes. At the same time, due to poor rendering and transmission loss, CGIs also suffer from low-level distortions such as blur and texture damage. Therefore, we propose a multi-stage feature fusion (MFF) module to make full use of the quality-aware information from low-level to high-level in CGIQA tasks as well.

Assume that there are $N_{s}$ stages, and $F_{i}$ is the extracted feature map from the $i$-th stage, where $i \in \{0,1, \ldots, N_{s}-1\}$. Considering that the feature maps obtained from different stages have different resolutions, we first apply the 2D adaptive average pooling with the target output size of 7$\times$7 to keep all feature maps having the same resolution:
\begin{equation}
    \hat{F_{i}} = A_{7\times7} F_{i}, 
\end{equation}
where $A_{7\times7}$ represents the adaptive average pooling operations with a target output size of 7$\times$7. Then the pooled feature maps are concatenated to form the combined feature maps $\hat{F} \in \mathbf{R}^{C \times 7 \times 7} $ ($C$ is the sum of the number of channels):
\begin{equation}
    \hat{F} = \mathop{\cup}\limits_{i=0}^{N_{s}-1} F_{i}, 
\end{equation}
where $\cup$ indicates the concatenation process. To dynamically explore the spatial information across different channels, we use a convolution block that consists of 3 convolution layers to generate the fused feature maps:
\begin{equation}
    \hat{F_{s}} = W_{1\times1}\times W_{3\times3}\times W_{1\times1} \times \hat{F},
\end{equation}
where $W_{1\times1}$ and $W_{3\times3}$ represent the 1$\times$1 and 3$\times$3 convolution operation. The first $1\times1$ convolution layer is used to reduce the channels to a quarter and generates feature maps $W_{1\times1} \times \hat{F} \in \mathbf{R}^{\frac{C}{4} \times 7 \times 7}$, which helps lower the computational complexity. Then the $3\times3$ convolution layer is applied to fuse the spatial information across channels and generates feature maps $ W_{3\times3} \times W_{1\times1} \times \hat{F} \in \mathbf{R}^{\frac{C}{4} \times 7 \times 7}$. Finally, the $1\times1$ convolution layer is employed to adjust the number of channels and we set the number of output channels as the same of the last stage's feature maps and we can get the fused feature maps $\hat{F_{s}} \in \mathbf{R}^{C' \times 7 \times 7}$ ($C'$ is the number of adjusted channels).

\subsubsection{Multi-stage Channel Attention}
\label{sec:MCA}
 Each channel of the feature maps is believed to function as a feature detector and focuses on representative meanings of the image \cite{woo2018cbam}. For example, the channels may represent specific aspects, possibly such as  
texture, color, illumination (low-level), and interactivity, understandability (high-level), etc, in CGIQA tasks. It is apparent that the features reflected by different channels have different impacts on human perception. Inspired by many channel attention works \cite{hu2018squeeze,li2019selective,woo2018cbam,park2018bam}, we introduce a typical channel attention module to improve the representation of the multi-stage fused feature maps as the multi-stage channel attention (MCA) module.  

At the beginning of this module, we simply squeeze the spatial dimension of the obtained feature maps $\hat{F_{s}}$ using both average-pooling and max-pooling. It is argued that average-pooling together with max-pooling can gather both global and distinctive features, which can help improve the representation power of channel-wise attention \cite{woo2018cbam}. The process can be derived as:
\begin{equation}
\begin{aligned}
    \hat{F_{a}} &= A_{1\times1} \hat{F_{s}}, \\
    \hat{F_{m}} &= M_{1\times1} \hat{F_{s}},
\end{aligned}
\end{equation}
where $A_{1\times1}$ and $M_{1\times1}$ stand for the average-pooling and max-pooling, and such squeeze process generates channel descriptors $\hat{F_{a}} \in \mathbf{R}^{C' \times 1 \times 1}$ and $\hat{F_{m}} \in \mathbf{R}^{C' \times 1 \times 1}$. Then the descriptors are put through multi-layer perceptron (MLP) with one hidden layer, the parameters of which are shared for $\hat{F_{a}}$ and $\hat{F_{m}}$. Similarly, to reduce the computation complexity, the hidden activation size of the MLP is set as $\mathbf{R}^{\frac{C'}{r} \times 1 \times 1}$, where $r$ is the reduction ratio. The MLP output size is still set the same as the input channel descriptors ($\mathbf{R}^{C' \times 1 \times 1}$). The channel attention feature maps can be computed as the sum of forwarded channel descriptors:
\begin{equation}
    \hat{F_{c}} = \delta (MLP(\hat{F_{a}}) \oplus MLP(\hat{F_{m}})),
\end{equation}
where $\oplus$ indicates the element-wise summation operation and $\delta$ denotes the sigmoid function. 
The final strengthened quality-wise feature vector can then be computed with the elementwise multiplication and average pooling:
\begin{equation}
    \mathbf{{F_D}} = Avg(\hat{F_{c}} \otimes \hat{F_{s}}),
\end{equation}
where $\otimes$ indicates the element-wise multiplication, $Avg(\cdot)$ represents the average pooling operation, the final distortion quality-wise feature vector $\mathbf{{F_D}} \in \mathbf{R}^{{C_D}\times1}$, the multi-stage channel attention feature maps $\hat{F_{c}} \in \mathbf{R}^{C' \times 1 \times 1}$, and multi-stage fused feature maps $\hat{F_{s}} \in \mathbf{R}^{C' \times 7 \times 7}$.

\subsection{Aesthetic Stream}
\label{sec:aes}
To specifically adapt to the AQA for CGIs and ensure evaluation robustness, we pre-train the backbone on the TGQA \cite{ling2020subjective} and AVA \cite{murray2012ava} databases. Considering that the aesthetic quality is highly correlated with the semantics \cite{li2023theme}, we only employ the high-level features (obtained from the last stage) for analysis. The backbone is first trained on the AVA database and then trained on the TGQA database since the AVA database is much larger in scale. The utilization of these two distinct databases and the sequential training process allows for the acquisition of both general and CGI-related aesthetic quality representations, contributing to a more comprehensive understanding of aesthetic perception in the context of the given task. Afterward, the aesthetic features can be derived as:
\begin{equation}
    \mathbf{{F_A}} = Avg(F_l),
\end{equation}
where $F_l$ is the feature map from the last stage of the backbone and the final aesthetic quality-wise feature vector $\mathbf{{F_A}} \in \mathbf{R}^{{C_A} \times 1}$. It's worth mentioning that the pre-trained weights of the aesthetic stream backbone are fixed to avoid the impact of distortion in the joint training. 

\subsection{Feature Fusion \& Regression}
With the extracted features, two Fully Connected (FC) layers are adopted as the feature regression module, which consists of 1024 and 128 neurons respectively:
\begin{equation}
    \hat{Q} = FC(MLP(\mathbf{{F_D}}) \cup MLP(\mathbf{{F_A}})),
\end{equation}
where $\hat{Q}$ represents the predicted quality score and the multi-layer perception is used to align the channels of the distortion and aesthetic features. The mean squared error (MSE) is utilized as the loss function:
\begin{equation}
    Loss = \frac{1}{m}\sum_{i=1}^{m}\left(\hat{Q}_{i}-Q_{i}\right)^{2},
\end{equation}
where $Q$ represents the ground-truth quality score obtained from subjective experiments and $m$ indicates the number of images in a mini batch.

\section{Experiment}
\label{sec:experiment}
In this section, we conduct experiments on the proposed CGIQA-6k and other CGIQA-related databases. Ablation experiments are carried out to demonstrate the contributions and effects of the proposed streams and modules. Cross-database validation is conducted to confirm the generalization ability of the proposed method. In-depth performance discussions are given as well.
\subsection{Comparing Models}
Since there are no pristine reference images in the proposed CGIQA-6k database, we only select NR IQA models for comparison. The chosen models can be divided into handcrafted-based models and deep learning-based models:
\begin{itemize}
    \item Handcrafted-based models  : These models include BRISQUE \cite{mittal2012no}, NIQE \cite{mittal2012making}, BIBLE \cite{li2015no}, BLIINDS2 \cite{saad2012blind}, BMPRI \cite{min2018blind}, CPBD \cite{narvekar2011no}, UCA \cite{min2017unified}, and NFERM \cite{gu2014using}. Such models extract handcrafted features through prior knowledge, and then employ the extracted features to evaluate the quality levels of images.
    \item Deep learning-based models: These models consist of DBCNN \cite{zhang2018blind}, MUSIQ \cite{ke2021musiq}, HyperIQA \cite{su2020blindly}, MGQA \cite{wang2021multi}, and StairIQA \cite{sun2021blind}. These mentioned models obtain quality-aware information by learning a feature representation from the labeled data and have achieved competitive performance on previous IQA tasks. 
\end{itemize}

\subsection{ Evaluation Criteria}
Four mainstream consistency evaluation criteria are used to measure the correlation between the predicted scores and MOS, which include Spearman Rank Correlation Coefficient (SRCC), Pearson Linear Correlation Coefficient (PLCC), Kendall’s Rank Correlation Coefficient (KRCC),  Root Mean Squared Error (RMSE). The SRCC values represent the similarity between two groups of rankings, the PLCC values describe the linear correlation of two sets of rankings, the KRCC values denote the ordinal association between two measured quantities, and the RMSE values stand for the average distance between the predicted scores and labels. 

Before obtaining the criteria values, a five-parameter logistic function is applied to map the predicted scores according to the practices in \cite{sheikh2006statistical}:
\begin{equation}
\hat{y}=\beta_{1}\left(0.5-\frac{1}{1+e^{\beta_{2}\left(y-\beta_{3}\right)}}\right)+\beta_{4} y+\beta_{5},
\end{equation}
where $\left\{\beta_{i} \mid i=1,2, \ldots, 5\right\}$ are parameters to be fitted, $y$ and $\hat{y}$ are the predicted scores and mapped scores respectively.
Besides, the MOSs for all databases are rescaled to a five-point scale for validation.

\begin{table*}[t]\scriptsize
\renewcommand\tabcolsep{2pt}
  \caption{Performance comparison on the CGIQA-6k database and the corresponding subsets. The best performance results are marked in {\bf\textcolor{red}{RED}} and the second performance results are marked in {\bf\textcolor{blue}{BLUE}}.}
  \vspace{-0.05cm}
  \begin{center}
  \begin{tabular}{c:l:l|cccc|cccc|cccc}
    \toprule
    \multirow{2}{*}{Index} & \multirow{2}{*}{Model} & \multirow{2}{*}{Type} & \multicolumn{4}{c|}{Game} & \multicolumn{4}{c|}{Movie} & \multicolumn{4}{c}{All}\\ \cline{4-15}
     &&& SRCC$\uparrow$ &  PLCC$\uparrow$ & KRCC$\uparrow$ & RMSE$\downarrow$ & SRCC$\uparrow$ &  PLCC$\uparrow$ & KRCC$\uparrow$ & RMSE$\downarrow$  & SRCC$\uparrow$ &  PLCC$\uparrow$ & KRCC$\uparrow$ & RMSE$\downarrow$\\
    \hline
    A&BRISQUE & \multirow{8}{30pt}{Hand-crafted-based} & 0.4231 & 0.4246 & 0.2855 & 0.5460 & 0.4434 & 0.4316 & 0.3083 & 0.5590  & 0.3979 & 0.4126 & 0.2735 & 0.6531  \\
    B&NIQE &  & 0.1185 & 0.1016 & 0.0907 & 0.6188 & 0.0326 & 0.0114 & 0.0245 & 0.6423  & 0.1091 & 0.0777 & 0.0821 & 0.7035  \\
    C&BIBLE &  & 0.0828 & 0.0704 & 0.0542 & 0.6265 & 0.0268 & 0.0334 & 0.0171 & 0.6448 & 0.0064 & 0.0077 & 0.0031 & 0.7025  \\
    D&BLIINDS2 &  & 0.3696 & 0.3509 & 0.2499 & 0.5872 & 0.3670  & 0.3693 & 0.2490 & 0.5927  & 0.3530 & 0.3449 & 0.2373 & 0.6704  \\
    E&BMPRI &  & 0.2305 & 0.2378 & 0.1548 & 0.6010 & 0.2245 & 0.2514 & 0.1536 & 0.6046  & 0.1927 & 0.1933 & 0.1292 & 0.6944  \\
    F&CPBD &  & 0.1952 & 0.1857 & 0.1307 & 0.6331 & 0.1643 & 0.1321 & 0.1119 & 0.6204  & 0.1519 & 0.1536 & 0.1014 & 0.7006  \\
    G&UCA &  & 0.3044 & 0.3211 & 0.2118 & 0.5822 & 0.3595 & 0.3807 & 0.2484 & 0.5613  & 0.2126 & 0.2254 & 0.1409 & 0.6921 \\
    H&NFERM &  & 0.3660 & 0.4241 & 0.2513 & 0.5543 & 0.3801 & 0.3897 & 0.2592 & 0.6035  & 0.3978 & 0.3679 & 0.2666 & 0.6747  \\ \hdashline
    I&DBCNN & \multirow{7}{30pt}{Deep learning-based} & 0.5907 & 0.6439 & 0.4423 & 0.5055 & 0.6533 & 0.6402 & 0.4668 & 0.5492  & 0.6437 & 0.6568 & 0.4593 & 0.5460   \\
    J&MUSIQ &  & 0.6067 & 0.6158 & 0.4347 & 0.5256 & 0.7181 & 0.7140 & 0.5293 & 0.4442  & 0.7104 & 0.7250 & 0.5251 & 0.4870 \\
    K&HypherIQA &  & 0.6271 & 0.6371 & 0.4485 & 0.5533 & 0.6465 & 0.6822 & 0.4666 & 0.4902  & 0.7003 & 0.7224 & 0.5160 & 0.5140 \\ 
    L&MGQA &  & \bf\textcolor{blue}{0.7472} & 0.7586 & \bf\textcolor{blue}{0.5579} & 0.3816 & \bf\textcolor{blue}{0.7494} & \bf\textcolor{blue}{0.7573} & \bf\textcolor{blue}{0.5648} & 0.3992 & 0.7999 & 0.8053 & 0.5865 & 0.4110\\
    M&StairIQA & & 0.7461 & \bf\textcolor{blue}{0.7601} & 0.5567 & \bf\textcolor{blue}{0.3806} & 0.7410 & 0.7493 & 0.5552 & \bf\textcolor{blue}{0.3938} & \bf\textcolor{blue}{0.8117} & \bf\textcolor{blue}{0.8186} & \bf\textcolor{blue}{0.6191} & \bf\textcolor{blue}{0.4082}\\
    N&Proposed &    & \bf\textcolor{red}{0.8169} & \bf\textcolor{red}{0.8209} & \bf\textcolor{red}{0.6431} & \bf\textcolor{red}{0.3783} & \bf\textcolor{red}{0.8234} & \bf\textcolor{red}{0.8303} & \bf\textcolor{red}{0.6577} & \bf\textcolor{red}{0.3865} & \bf\textcolor{red}{0.8552} & \bf\textcolor{red}{0.8531} & \bf\textcolor{red}{0.6444} & \bf\textcolor{red}{0.3781}\\
    
    \bottomrule
  \end{tabular}
  \end{center}
  \label{tab:cgiqa}

\end{table*}

\begin{figure*}[t]
    \centering
    \includegraphics[width=\linewidth]{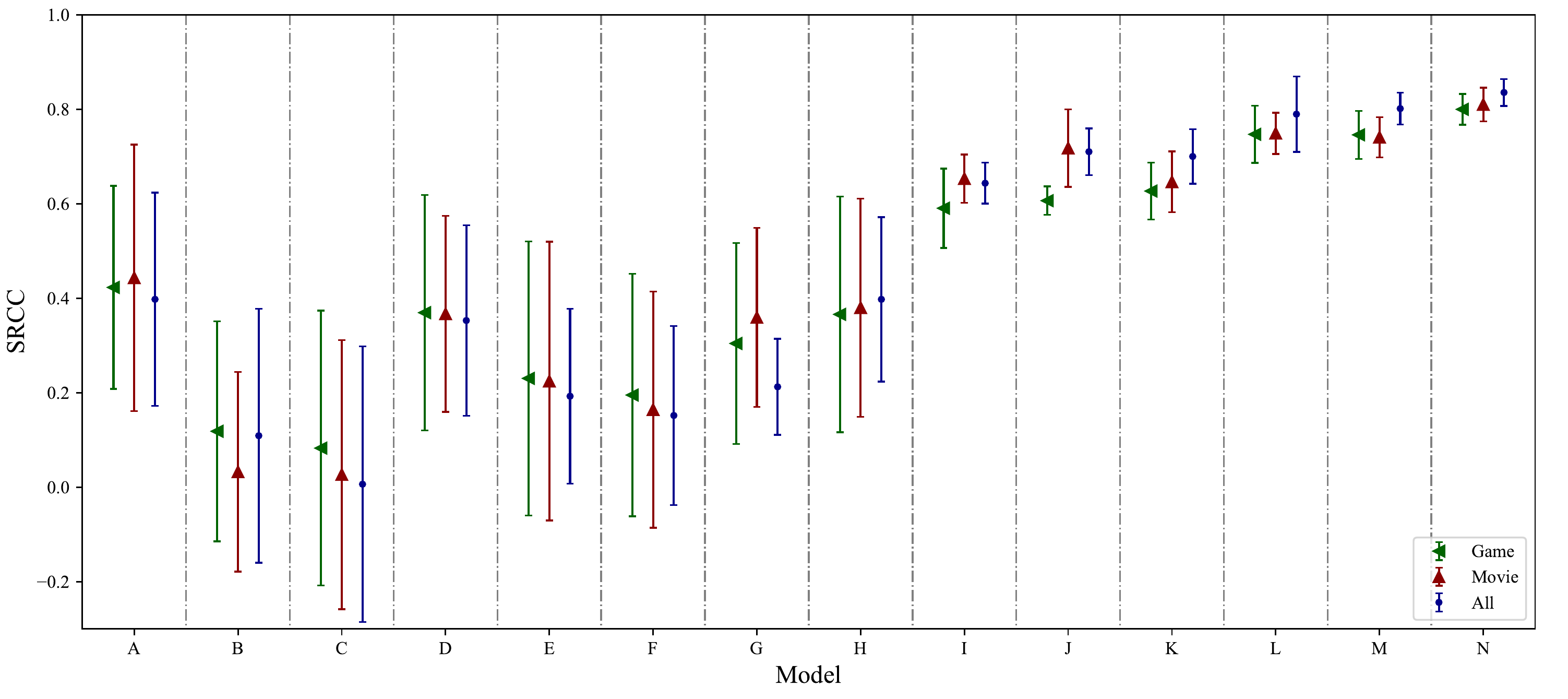}
    \caption{Mean SRCC values and standard error bars for methods compared on the CGIQA-6k database and its corresponding subsets. A-N are of the same order as in Table \ref{tab:cgiqa}. }
    \label{fig:errorbar}
\end{figure*}

\subsection{Experimental Setup}
\subsubsection{Aesthetic Pre-training Setup}
\begin{itemize}
    \item Input Size \& Backbone: The 224$\times$224 views are adopted as inputs by random-cropping the resized images with dimensions of 256$\times$256. The Swin Transformer tiny (ST-t) \cite{liu2021swin} is initialized with the weights pretrained on the ImageNet-22K database \cite{deng2009imagenet}. 
    \item Optimizer: The Adam optimizer \cite{kingma2014adam} is employed with an initial learning rate set as 1e-5. The default batch size is set as 32. 
    \item Pre-training Databases: The AVA \cite{murray2012ava} database, which includes 250,000 aesthetic images, and the TGQA \cite{ling2020subjective} database, which includes 1091 aesthetic mobile game images, are used as the pre-training databases. To begin, the aesthetic stream is trained on the AVA database for 50 epochs. This initial training is aimed to acquire a comprehensive understanding of general aesthetic quality representations. Following this, the aesthetic stream is further trained on the TGQA database for an additional 10 epochs. This subsequent training stage focuses on enhancing the aesthetic stream's ability to capture aesthetic quality representations specific to CGIs.
\end{itemize}

\subsubsection{Main Experimental Setup}
The details of the experimental setup are shown as follows:
\begin{itemize}
    \item Input Size \& Backbone: Similarly, the 224$\times$224 views and the ImageNet-22K database \cite{deng2009imagenet} initialized Swin Transformer tiny (ST-t) \cite{liu2021swin} backbone are utilized. Specifically, the reduction ratio $r$, the channel number $C_D$, and the channel number $C_A$ mentioned in Section \ref{sec:MFF}, \ref{sec:MCA}, and \ref{sec:aes} are 16, 720, and 768 respectively. For other comparing deep learning-based methods, the default training parameters are utilized.
    \item Optimizer: The Adam optimizer \cite{kingma2014adam} is employed with an initial learning rate set as 1e-5. The default batch size is set as 32. Each training process lasts for 40 epochs.
    \item Training and Testing set: We randomly separate the databases into the training sets and testing sets with a ratio of 8:2. We repeat the training process 10 times and the average results are recorded as the final performance.
\end{itemize}


\begin{table*}[t]\scriptsize
\renewcommand\tabcolsep{2pt}
  \caption{Performance results on the CCT-CGI, NBU-CIQAD, and LIVE-YT-Gaming databases. The best performance results are marked in {\bf\textcolor{red}{RED}} and the second performance results are marked in {\bf\textcolor{blue}{BLUE}}.}
  \vspace{-0.05cm}
  \begin{center}
  \begin{tabular}{c:l:l|cccc|cccc|cccc}
    \toprule
    \multirow{2}{*}{Index} & \multirow{2}{*}{Model}  & \multirow{2}{*}{Type} & \multicolumn{4}{c|}{CCT-CGI} & \multicolumn{4}{c|}{NBU-CIQAD} & \multicolumn{4}{c}{LIVE-YT-Gaming} \\ \cline{4-15}
     &&& SRCC$\uparrow$ &  PLCC$\uparrow$ & KRCC$\uparrow$ & RMSE$\downarrow$ & SRCC$\uparrow$ &  PLCC$\uparrow$ & KRCC$\uparrow$ & RMSE$\downarrow$  & SRCC$\uparrow$ &  PLCC$\uparrow$ & KRCC$\uparrow$ & RMSE$\downarrow$ \\
    \hline
    A&BRISQUE  & \multirow{8}{30pt}{Hand-crafted-based} & 0.8468 & 0.8609 & 0.6633 & 0.3605 & 0.4858 & 0.4579 & 0.3744 & 0.6860 & 0.4815 & 0.4986 & 0.3397 & 0.7437  \\
    B&NIQE  & & 0.6888 & 0.6862 & 0.5078 & 0.6760 & 0.4932 & 0.4985 & 0.3412 & 0.8329 & 0.5412 & 0.5864 & 0.3852 & 0.6990\\
    C&BIBLE  & & 0.5941 & 0.6272 & 0.4195 & 0.7204 & 0.1452 & 0.1071 & 0.0950 & 1.0008 &  0.1647 & 0.1344 & 0.1108 & 0.9782 \\
    D&BLIINDS2 & & 0.8894 & 0.8790 & 0.7082 & 0.3124 & 0.3915 & 0.3825 & 0.2700 & 0.9216 & 0.4204 & 0.4839 & 0.2915 &  0.7294  \\
    E&BMPRI  & & 0.8700 & 0.8770 & 0.6751 & 0.3292 & 0.4001 & 0.4140 & 0.2703 & 0.9258 & 0.4461 & 0.4844 & 0.3118 & 0.7746 \\
    F&CPBD  & & 0.6612 & 0.6490 & 0.4799 & 0.4579 & 0.2295 & 0.2414 & 0.1860 & 0.8958 & 0.0184 & 0.0422 & 0.0112 & 0.8232 \\
    G&UCA  & & 0.8459 & 0.8854 & 0.6450 & 0.5822 & 0.2916 & 0.1642 & 0.0817 & 0.9554 & 0.2503 & 0.2910 & 0.1734 & 0.8891 \\
    H&NFERM  & & 0.3570 & 0.3340 & 0.2413 & 0.6094 & 0.0848 & 0.0736 & 0.0580 & 1.0048 & 0.2035 &0.1684 & 0.1385 &0.8782 \\\hline
    I&DBCNN  & \multirow{7}{30pt}{Deep learning-based} & 0.9111 & 0.9123 & 0.7365 & 0.2911 & 0.7295 & 0.7301 & 0.5296 & 0.7162  & 0.6832 & 0.5240 & 0.5035 & 0.7989 \\
    J&MUSIQ  & & 0.9077 & 0.9045 & 0.7481 & 0.2710 & 0.8419 & 0.8531 & 0.6481 & 0.5792 & 0.6107 & 0.4095 & 0.4435 & 0.8094\\
    K&HypherIQA  && 0.9122 & 0.9219 & 0.7437 & 0.2699 & 0.6297 & 0.6771 & 0.5383 & 0.5831 & 0.5200 & 0.5840 & 0.3705 & 0.7183    \\ 
    
    L&MGQA  && 0.9061 & 0.9062 & \bf\textcolor{blue}{0.7825} & 0.2901 & 0.8916 & 0.8842 & 0.7113 & 0.4712  & \bf\textcolor{blue}{0.7266} & \bf\textcolor{blue}{0.7791} & \bf\textcolor{blue}{0.5376} & \bf\textcolor{blue}{0.5312}\\
    M&StairIQA  && \bf\textcolor{blue}{0.9163} & \bf\textcolor{blue}{0.9142} & {0.7660} & \bf\textcolor{blue}{0.2454} & \bf\textcolor{blue}{0.9070} & \bf\textcolor{blue}{0.9041} & \bf\textcolor{blue}{0.7277} & \bf\textcolor{blue}{0.4311}  & 0.6808 & 0.7024 & 0.4903 & 0.6032\\
    N&Proposed   && \bf\textcolor{red}{0.9210} & \bf\textcolor{red}{0.9230} & \bf\textcolor{red}{0.7851} & \bf\textcolor{red}{0.2220} & \bf\textcolor{red}{0.9335} & \bf\textcolor{red}{0.9332} & \bf\textcolor{red}{0.7838} & \bf\textcolor{red}{0.3547} & \bf\textcolor{red}{0.7613} & \bf\textcolor{red}{0.7733} & \bf\textcolor{red}{0.5612} & \bf\textcolor{red}{0.5212}\\
    
    \bottomrule
  \end{tabular}
  \end{center}
  \label{tab:others}
  \vspace{-0.3cm}
\end{table*}


\subsection{Performance on the CGIQA-6k Database}
The summary of experimental performance on the CGIQA-6k database and its corresponding subsets is presented in Table \ref{tab:cgiqa}. The table highlights the best performance achieved for each column. Additionally, to further validate the effectiveness and stability of various IQA methods, the mean SRCC values and corresponding standard error bars are illustrated in Fig. \ref{fig:errorbar}. The following observations can be made based on the aforementioned results:
a) Deep learning-based methods demonstrate significant advantages over handcrafted-based methods in terms of both mean and standard deviation values. This disparity arises from the fact that handcrafted-based methods are designed to extract features suited for natural scene content rather than computer graphics content. Conversely, deep learning-based methods have the capability to learn more effective feature representations specifically tailored for CGIs.
b) With the exception of DBCNN \cite{zhang2018blind} and MUSIQ \cite{ke2021musiq}, most deep learning-based methods exhibit improved performance across the entire CGIQA-6k database compared to the separate subsets. This observation indicates that learning from a larger sample size contributes to a better understanding of CGIs.
c) All deep learning-based methods achieve better performance on the movie subset compared to the game subset. This can be attributed to the movie CGIs being more refined, thereby making the perceived distortion levels more discernible. Conversely, the game CGIs tend to have lower initial quality levels due to limitations in computing resources, resulting in relatively inconspicuous distortions.
d) The proposed method demonstrates the highest performance on the CGIQA-6k database, including both subsets, and exhibits the smallest standard deviation values among all IQA methods considered in the comparison. This suggests that the proposed method performs more effectively and consistently.

\subsection{Performance on other CGIQA-related Databases}
\subsubsection{Databases Selection}
To further demonstrate the effectiveness of the proposed method, we also select several existing CGIQA-related databases to compare the performance of the proposed model and the mainstream IQA models, which include the CCT \cite{min2017unified},
NBU-CIQAD \cite{chen2021perceptual}, LIVE-YT-Gaming \cite{yu2022subjective}. A detailed introduction of the selected databases is given as follows:
\begin{itemize}
    \item CCT: The CCT database is a cross-content-type IQA database that deals with the quality assessment for natural scene images (NSIs), computer graphic images (CGIs), and screen content images (SCIs). We only adopt the 528 distorted CGIs for validation and such subset of the CCT database is referred to as the CCT-CGI database in this paper. The distortions are generated by HEVC \cite{sullivan2012overview} and HEVC-SCC \cite{xu2015overview} standards.
    \item NBU-CIQAD: The NBU-CIQAD database focuses on the quality assessment of cartoon images, which has a high correlation with CGIs. The NBU-CIQAD database consists of  1,800 cartoon images corrupted by single type of distortion and 800 cartoon images suffering from multiple types of distortions. The distortions are caused by the manual adjustment of brightness, saturation, and contrast. 
    \item LIVE-YT-Gaming: The LIVE-YT-Gaming database includes 600 authentic user-generated content (UGC) gaming videos and 18,600 subjective quality ratings collected from an online subjective study. We extract frames from the gaming videos with a fixed interval of 60 frames and obtain 3,501 game images for validation. The images share the same quality scores as the corresponding videos.
    
\end{itemize}
We maintain the default experimental setup. Additionally, the distorted images of the CCT-CGI and the NBU-CIQAD databases are synthesized from the corresponding reference images. Therefore, we separate the training and testing sets without content overlap.
\subsubsection{Performance discussion}
Table \ref{tab:others} presents the performance of the proposed method on other CGIQA-related IQA databases. A detailed analysis of the results is provided below:
a) The CCT-CGI database utilizes manual compression distortions for evaluation. Both handcrafted-based and deep learning-based methods demonstrate relatively good performance. Notably, all deep learning-based methods achieve SRCC values higher than 0.9.
b) The NBU-CIQAD database focuses specifically on color distortions. In this task, deep learning-based methods outperform handcrafted-based methods. This can be attributed to the fact that handcrafted features have a low correlation with color information, making deep learning approaches more effective in this context.
c) The LIVE-YT-Gaming database consists of authentic distorted contents. As a result, all IQA models experience a noticeable performance drop compared to other databases. Assessing authentic distortions is considered more challenging, which contributes to the lower performance across the models.
These observations shed light on the strengths of the proposed method in relation to other CGIQA-related IQA databases, showcasing its performance in different distortion scenarios.

\begin{table*}[t]\scriptsize
\caption{Performance Results of the ablation study on the two streams. The mark \checkmark indicates that the referred stream  is applied while $\times$ indicates that the referred stream is not applied. The aesthetic stream backbone weights are fixed and only linear regression is employed for the single aesthetic stream.}
    \centering
    \begin{tabular}{c:c|cc|cc|cc|cc}
    \toprule
         \multirow{2}{*}{Distortion}   &\multirow{2}{*}{Aesthetic}   &\multicolumn{2}{c|}{CGIQA-6K} &\multicolumn{2}{c|}{CCT-CGI} & \multicolumn{2}{c|}{NBU-CIQAD} &\multicolumn{2}{c}{LIVE-YT}   \\ \cline{3-10}
         && SRCC$\uparrow$ & PLCC$\uparrow$  & SRCC$\uparrow$ & PLCC$\uparrow$ & SRCC$\uparrow$ & PLCC$\uparrow$ & SRCC$\uparrow$ & PLCC$\uparrow$ \\ \hline
         \checkmark  & \checkmark     & \bf\textcolor{red}{0.8552} & \bf\textcolor{red}{0.8531} & \bf\textcolor{red}{0.9210} & \bf\textcolor{red}{0.9230} & \bf\textcolor{red}{0.9335} & \bf\textcolor{red}{0.9332} & \bf\textcolor{red}{0.7613} & \bf\textcolor{red}{0.7733} \\
        \checkmark  & $\times$        & \bf\textcolor{blue}{0.8288}  & \bf\textcolor{blue}{0.8353} & \bf\textcolor{blue}{0.9111} & \bf\textcolor{blue}{0.9033} & \bf\textcolor{blue}{0.9297} & \bf\textcolor{blue}{0.9292} & \bf\textcolor{blue}{0.7451} & \bf\textcolor{blue}{0.7471}\\
       $\times$   & \checkmark        & 0.3002  & 0.3091 & 0.4456 &0.4245 & 0.5345 & 0.5375 &0.3801 & 0.3834 \\  
    \bottomrule
    \end{tabular}
    \label{tab:ablation_stream}
    \vspace{-0.3cm}
\end{table*}

\begin{table*}[t]\scriptsize
\caption{Performance Results of the ablation study on the attention modules. The mark \checkmark indicates that the referred module is applied while $\times$ indicates that the referred module is not applied.}
    \centering
    \begin{tabular}{c:c|cc|cc|cc|cc}
    \toprule
         \multirow{2}{*}{MFF}   &\multirow{2}{*}{MCA}   &\multicolumn{2}{c|}{CGIQA-6K} &\multicolumn{2}{c|}{CCT-CGI} & \multicolumn{2}{c|}{NBU-CIQAD} &\multicolumn{2}{c}{LIVE-YT}   \\ \cline{3-10}
         && SRCC$\uparrow$ & PLCC$\uparrow$  & SRCC$\uparrow$ & PLCC$\uparrow$ & SRCC$\uparrow$ & PLCC$\uparrow$ & SRCC$\uparrow$ & PLCC$\uparrow$ \\ \hline
        \checkmark  & \checkmark     & \bf\textcolor{red}{0.8552} & \bf\textcolor{red}{0.8531} & \bf\textcolor{red}{0.9210} & \bf\textcolor{red}{0.9230} & \bf\textcolor{red}{0.9335} & \bf\textcolor{red}{0.9332} & \bf\textcolor{red}{0.7613} & \bf\textcolor{red}{0.7733} \\
    \checkmark  & $\times$        & 0.8399  & 0.8316  &0.9107 &0.9122 & 0.9201 & 0.9214 & 0.7331 & 0.7343\\
     $\times$   & \checkmark        & \bf\textcolor{blue}{0.8410}  & \bf\textcolor{blue}{0.8427} & \bf\textcolor{blue}{0.9200} &\bf\textcolor{blue}{0.9119}  & \bf\textcolor{blue}{0.9276} & \bf\textcolor{blue}{0.9264} & \bf\textcolor{blue}{0.7467} & \bf\textcolor{blue}{0.7481}\\  
     $\times$   &  $\times$       & 0.8310  & 0.8368 &0.9077 &0.9086 & 0.9174 &0.9163 & 0.7300 &0.7321\\
    \bottomrule
    \end{tabular}
    \label{tab:ablation_distortion}
    \vspace{-0.3cm}
\end{table*}

\subsection{Ablation Study}

\subsubsection{Two Streams Contribution}
In this section, we investigate the contributions of the distortion and aesthetic streams employed in the proposed model for evaluating the visual quality of CGIs. The results of the ablation study for the two streams are presented in Table \ref{tab:ablation_stream}, revealing the following findings:
Firstly, employing both streams together yields better performance compared to using each single stream separately. This observation confirms the contributions of both the distortion and aesthetic streams in evaluating CGI quality, emphasizing the importance of considering both aspects.
Secondly, the aesthetic stream demonstrates significantly inferior performance compared to the distortion stream. This finding suggests that distortions have a more substantial impact on human perception when it comes to CGIs. It indicates that the presence of distortions plays a more critical role in shaping the perceived quality of CGIs, while the aesthetic attributes have a relatively less pronounced influence.

\subsubsection{Distortion Stream Modules Contribution}
To quantify the contributions of each stream in the proposed method, we conduct an ablation study while maintaining the default experiment setup. The results of the ablation study are presented in Table \ref{tab:ablation_distortion}, providing several notable conclusions.
Firstly, when the MFF and MCA modules are separately attached, they outperform the bare backbone model. This suggests that both of the proposed modules contribute to the final results, indicating their individual effectiveness.
Secondly, employing both the MFF and MCA modules together achieves the highest performance, further confirming the effectiveness of the proposed framework as a whole.
Thirdly, employing only the MFF module results in lower performance compared to employing only the MCA module. This finding suggests that the MCA module seems to make a more significant contribution to the proposed framework. It implies that channel attention plays a crucial role, and human perception of CGIs is more easily influenced by specific feature attributes.


\begin{table}[t]\scriptsize
\caption{{Performance results of the different backbones on the CGIQA-6k database. The proposed framework is adaptively applied to the backbones. The number of parameters and flops are exhibited as well.}}
    \centering
    \begin{tabular}{l|cccccc}
    \toprule
         Backbone      & SRCC$\uparrow$    & PLCC$\uparrow$   & KRCC$\uparrow$   & RMSE$\downarrow$  & FLOPs$\downarrow$ & Params$\downarrow$\\ \hline
        VGG16   & 0.8212  & 0.8219 & 0.6087 & 0.4254 &30.94G &277.74M\\
        ResNet50  & 0.8312  & 0.8306 & 0.6111 & 0.4114 &\bf\textcolor{blue}{8.27G} & \bf\textcolor{blue}{52.14M}\\ 
        MobileNetV2    & 0.7875  & 0.7918 & 0.5855 & 0.4339 &\bf\textcolor{red}{0.66G} & \bf\textcolor{red}{8.03M}\\ 
        ConvNeXt-tiny    & \bf\textcolor{blue}{0.8419}  & \bf\textcolor{blue}{0.8481} & \bf\textcolor{blue}{0.6551} & \bf\textcolor{blue}{0.3768} & 8.91G &55.99M\\
        Swin-tiny &\bf\textcolor{red}{0.8552} & \bf\textcolor{red}{0.8531} & \bf\textcolor{red}{0.6444} & \bf\textcolor{red}{0.3781} &8.94G &59.92M\\
    \bottomrule
    \end{tabular}
    \vspace{-0.25cm}
    \label{tab:backbone}
\end{table}

\subsection{Backbone Comparison}
The proposed framework can be easily adapted to other CNN modules which contain multiple stages. To further test the performance of different backbones, we select VGG16 \cite{simonyan2014very}, ResNet50 \cite{he2016deep}, MobileNetV2 \cite{sandler2018mobilenetv2}, and ConvNeXt-tiny \cite{liu2022convnet} for comparison. The performance results on the CGIQA-6k database are clearly exhibited in Table \ref{tab:backbone}, from which we can make several useful observations. First, the effectiveness of the proposed framework is demonstrated by the competitive performance achieved by all backbones in comparison to other NR IQA methods. This outcome highlights the capability of the proposed framework to effectively address the task at hand.
Second, the Swin Transformer tiny backbone achieves the best performance among the evaluated models. This finding suggests that the Swin Transformer tiny architecture is particularly well-suited for the specific requirements of the task.
Additionally, the light-weight MobileNetV2 model also demonstrates strong performance, indicating its potential for deployment on mobile devices by only consuming 0.66G flops and 8.03M parameters. This finding highlights the versatility of the proposed framework, which can accommodate various popular CNN models, thus enabling broader applications in CGIQA tasks with diverse demands.

\begin{table}[t]\scriptsize
\caption{Experimental results for the cross-database validation. The NR IQA models are trained on the CGIQA-6K database. The default experimental setup is maintained.}
    \centering
    \begin{tabular}{l|cc|cc}
    \toprule
         \multirow{2}{*}{Model}  &\multicolumn{2}{c|}{CCT-CGI} & \multicolumn{2}{c}{NBU-CIQAD}  \\ \cline{2-5}
         & SRCC$\uparrow$ & PLCC$\uparrow$  & SRCC$\uparrow$ & PLCC$\uparrow$ \\ \hline
      MGQA & 0.7088 & 0.7086 & \bf\textcolor{blue}{0.2341} & \bf\textcolor{blue}{0.2797}\\
      StairIQA& \bf\textcolor{blue}{0.7211}  & \bf\textcolor{blue}{0.7297} & 0.2275 & 0.2608 \\
      Proposed& \bf\textcolor{red}{0.8092}  & \bf\textcolor{red}{0.8293} & \bf\textcolor{red}{0.3061} & \bf\textcolor{red}{0.3003} \\  
    \bottomrule
    \end{tabular}
    \label{tab:cross}
    \vspace{-0.25cm}
\end{table}

\subsection{Cross-Database Validation}
In this section, we perform cross-database validation to demonstrate the robustness and generalization ability of the proposed method. We select the most competitive NR IQA models, namely MGQA and StairIQA, and train all models on the CGIQA-6K database. Subsequently, we evaluate the models on the CCT-CGI and NBU-CIQAD databases (excluding the LIVE-YT-Gaming database, which focuses on videos).
The validation results are presented in Table \ref{tab:cross}, leading to the following observations:
a) The proposed method outperforms all the competitors by a significant margin, indicating its strong generalization ability. This suggests that the proposed method is capable of performing well on unseen databases, highlighting its robustness and effectiveness in assessing CGI quality.
b) The cross-database validation results on the NBU-CIQAD database are relatively unsatisfactory, and we provide insights into the reasons behind this performance gap. The CGIQA-6K database primarily consists of CGIs and encompasses a wide range of in-the-wild distortions. In contrast, the NBU-CIQAD database consists solely of cartoon images and primarily focuses on color-specific distortions. This significant disparity in content and distortions between the CGIQA-6K and NBU-CIQAD databases leads to poor performance. However, the CCT-CGI database comprises CGIs and primarily focuses on compression distortion, which falls within the scope of in-the-wild distortions. Consequently, the cross-database validation results on the CCT-CGI database are more impressive.

\begin{figure*}[t]
    \centering
    \subfigure[Game]{\includegraphics[width = .3\linewidth]{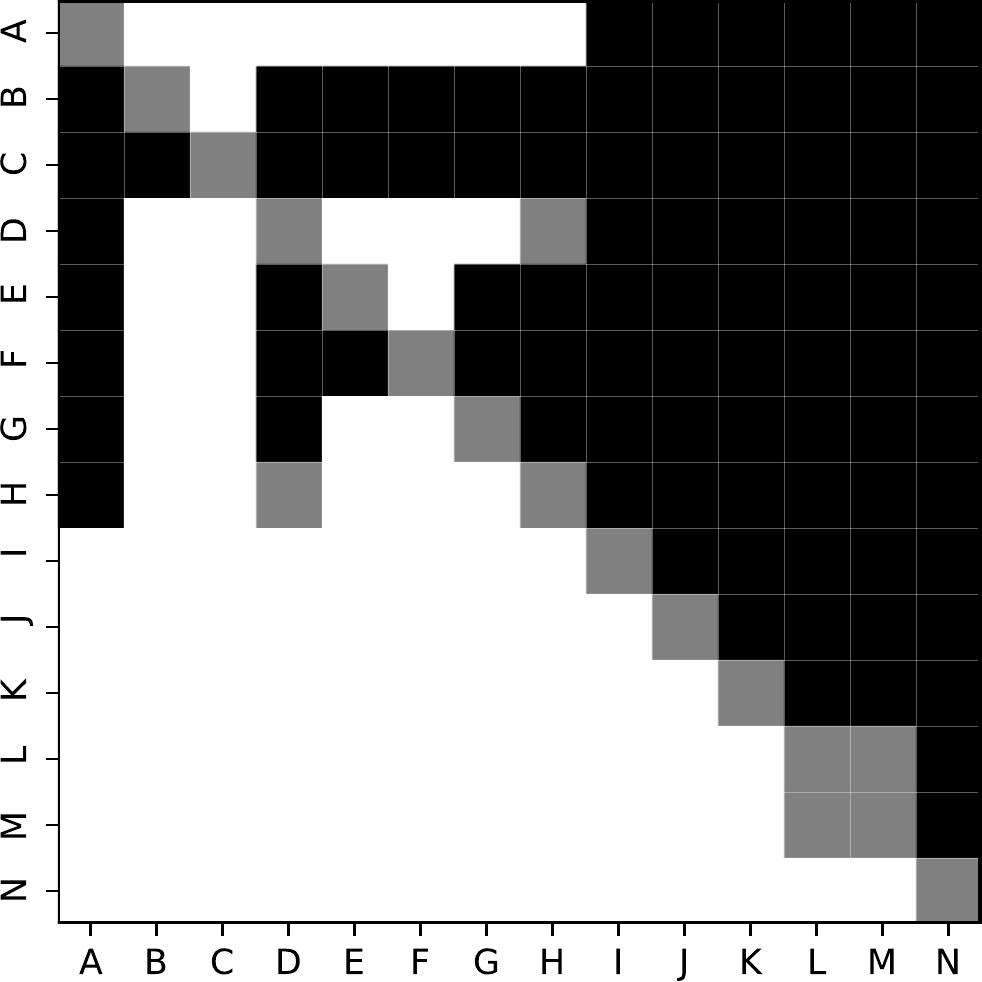}} 
    \subfigure[Movie]{\includegraphics[width = .3\linewidth]{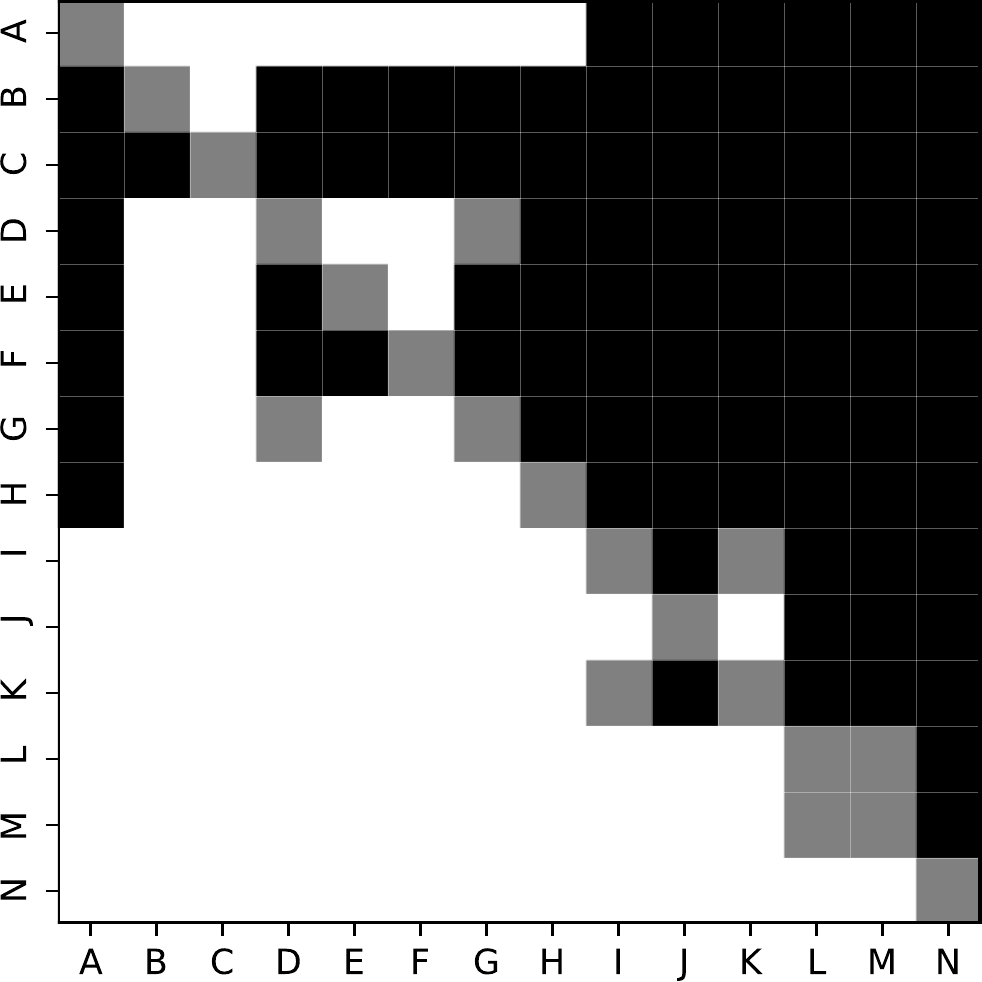}}
    \subfigure[All]{\includegraphics[width = .3\linewidth]{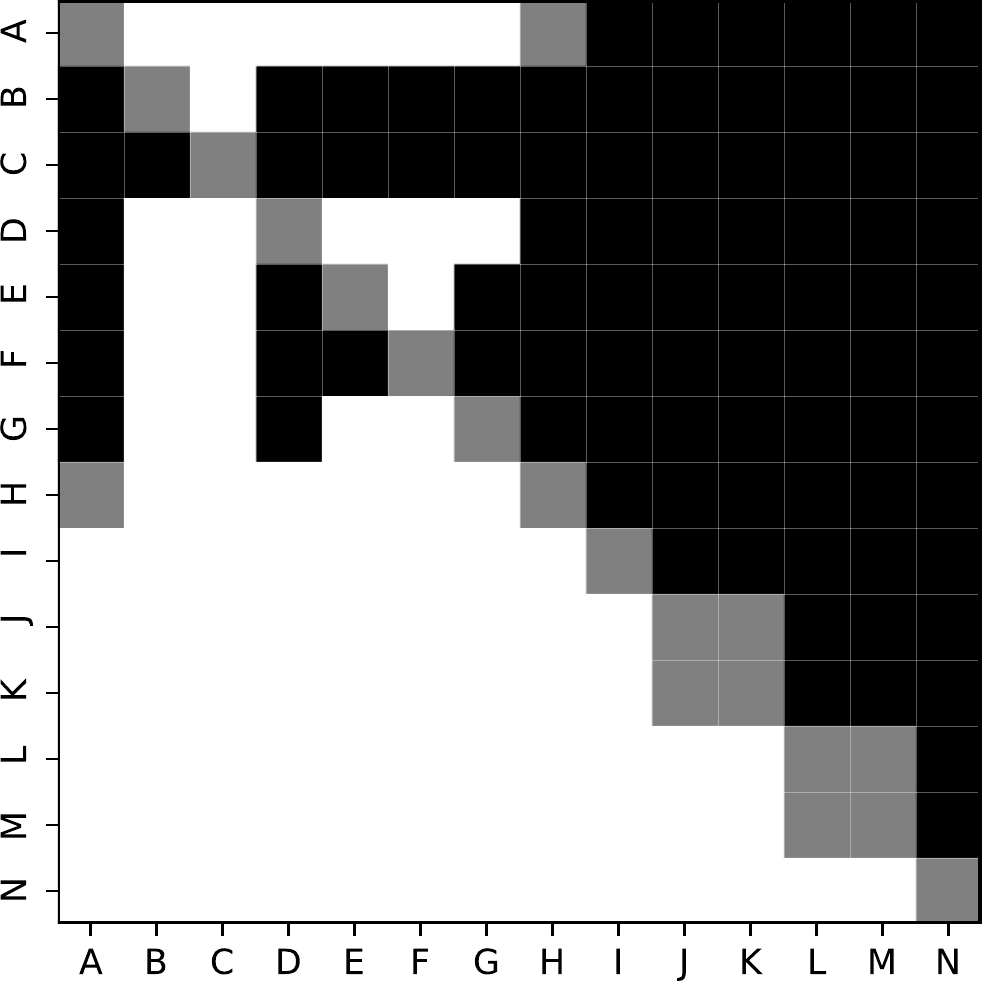}}
    \subfigure[CCT-CGI]{\includegraphics[width = .3\linewidth]{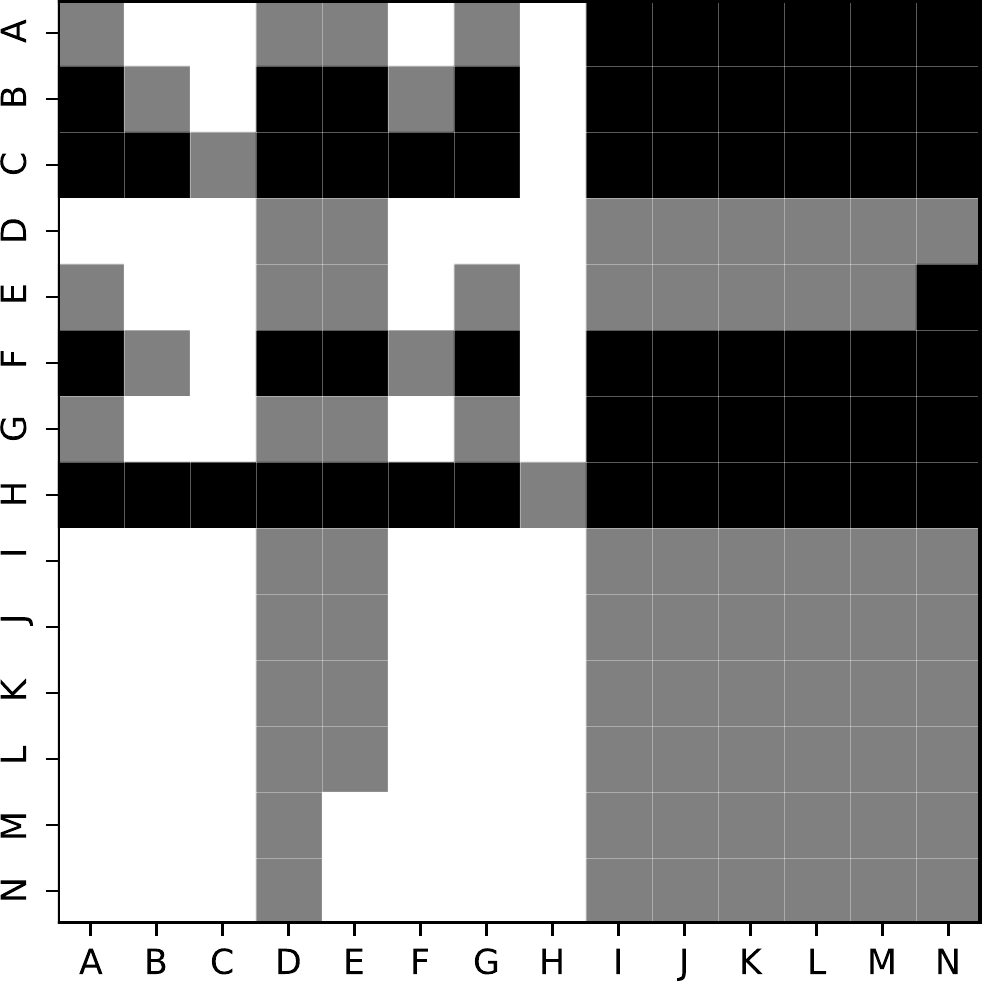}} 
    \subfigure[NBU-CIQAD]{\includegraphics[width = .3\linewidth]{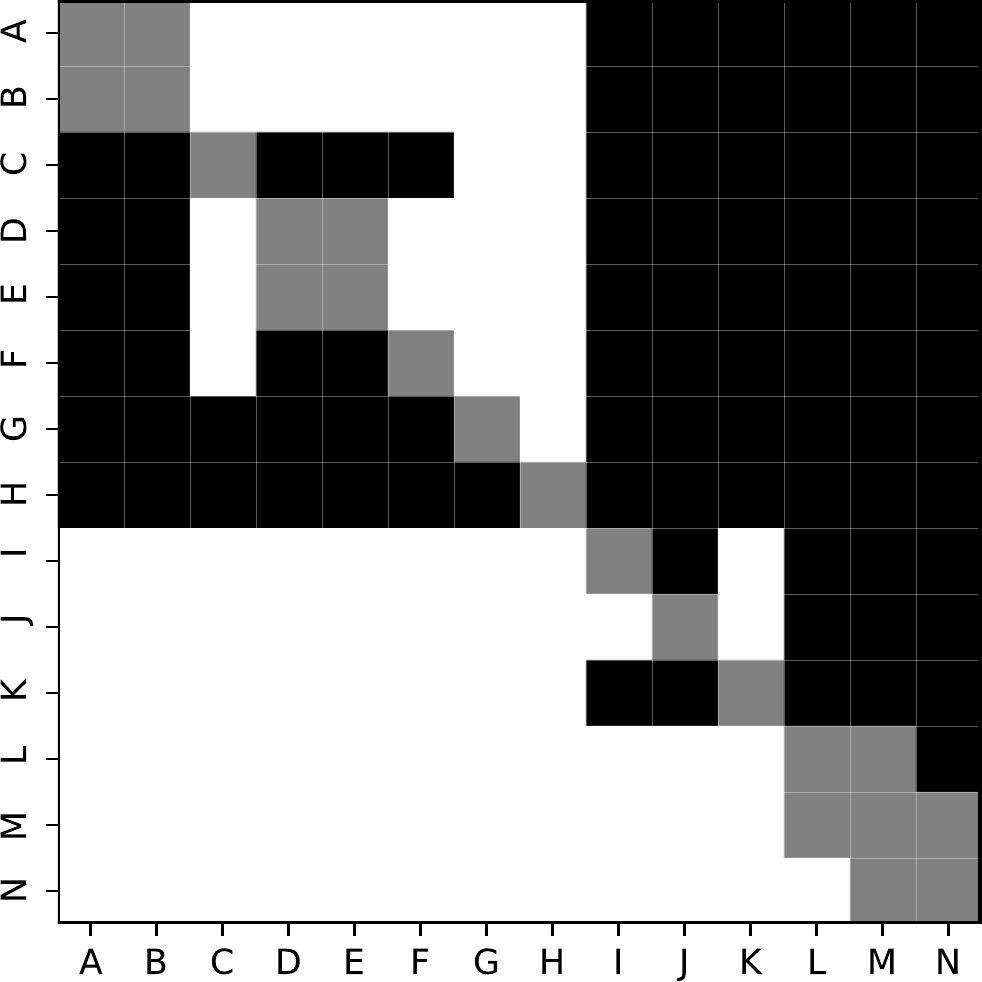}}  
    \subfigure[LIVE-YT-Gaming]{\includegraphics[width = .3\linewidth]{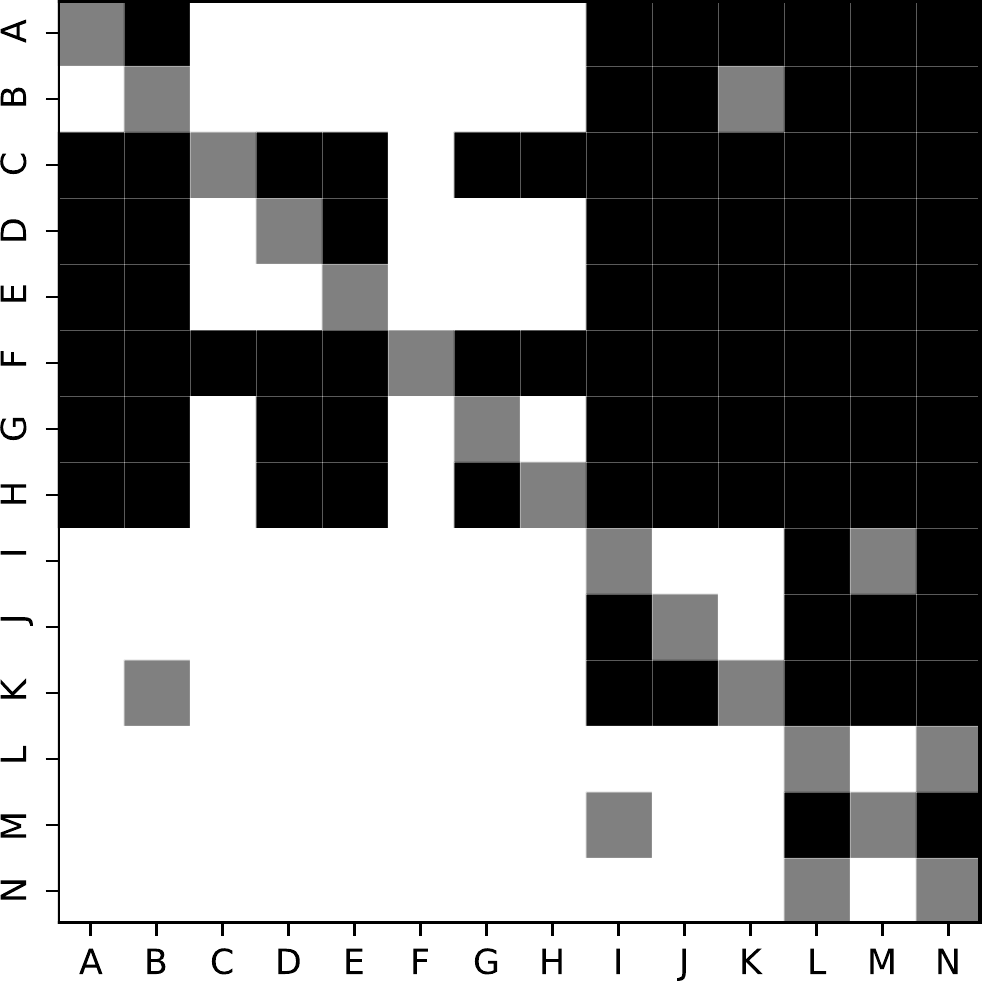}}
    \caption{Statistical test results on the proposed CGIQA-6k, CCT-CGI, NBU-CIQAD, and LIVE-YT-Gaming databases. A black/white block means the row method is statistically worse/better than the column one. A gray block means the row method and the column method are statistically indistinguishable. The metrics are denoted by the same index as in Table \ref{tab:cgiqa} and Table \ref{tab:others} respectively.}
    \label{fig:heatmap}
\end{figure*}

\subsection{Statistical Test}
{
In this section, we conduct a statistical test to further validate the effectiveness of the proposed method. We adhere to the experimental setup outlined in \cite{statistic-test} and compare the disparity between the predicted quality scores and the subjective ratings. The null hypothesis posits that the residuals from one quality assessment model originate from the same distribution and are, at a 95\% confidence level, statistically indistinguishable from the residuals of another quality assessment model. All possible pairs of models are examined, and the outcomes are presented in Fig. \ref{fig:heatmap}.}
Upon closer examination, we observe that the proposed method exhibits statistical superiority over all the compared NR IQA models on the CGIQA-6k database and its corresponding subsets. Additionally, our method demonstrates competitive performance on other CGIQA-related databases. Specifically, the proposed method statistically outperforms 6, 12, and 13 NR IQA models on the CCT-CGI, NBU-CIQAD, and LIVE-YT-Gaming databases, respectively.
These findings reinforce the efficacy of our proposed method, establishing its significant statistical advantage in the domain of CGIQA and its associated databases.

\section{Conclusion}
\label{sec:conclusion}
In this paper, we construct a large-scale in-the-wild CGIQA database named CGIQA-6k. The database consists of 3,000 game CGIs and 3,000 movie CGIs and covers a wider range of resolutions. Through meticulous subjective experiments conducted in a controlled laboratory environment, we have obtained accurate perceptual ratings for the CGIs in the database. This database serves as a valuable resource for advancing research in the field.
Furthermore, we propose an effective deep learning-based NR IQA model that leverages both distortion and aesthetic quality representation. Our model outperforms all other state-of-the-art NR IQA methods on the CGIQA-6k database as well as other CGIQA-related databases. The superior performance demonstrates the efficacy and robustness of our approach in assessing CGI quality.
In summary, this work contributes to bridging the gap between evaluating NSIs and CGIs by constructing a comprehensive CGIQA database and proposing a powerful deep learning-based NR IQA model. The availability of the CGIQA-6k database will facilitate further research and advancements in the field of CGIQA.

\bibliographystyle{ACM-Reference-Format}
\bibliography{sample-base}










\end{document}